%% file: main.tex
\documentclass{cambrian}

\usepackage[authoryear,round]{natbib}
\setcitestyle{authoryear,round,citesep={;},aysep={,},yysep={;}}

\definecolor{logoindigo}{HTML}{3F3E7B}
\definecolor{darkblue}{rgb}{0, 0, 0.5}
\hypersetup{breaklinks=true,colorlinks=true,allcolors=logoindigo}
\usepackage{placeins}
\usepackage{afterpage}

\usepackage{thmtools}
\usepackage[normalem]{ulem}
\usepackage{siunitx}
\usepackage{float}
\usepackage{multirow}
\usepackage{makecell}
\usepackage{subfig}
\usepackage{wrapfig}
\usepackage{pdfpages}
\usepackage{soul}
\usepackage[most]{tcolorbox}
\usepackage{fontawesome5}
\usepackage{capt-of}
\usepackage{etoolbox}
\usepackage{adjustbox}
\usepackage{threeparttable}
\usepackage{pgfplots}
\pgfplotsset{compat=1.18}

\input{math_commands.tex}

\definecolor{deemph}{gray}{0.6}
\definecolor{groupgray}{gray}{0.90}
\definecolor{probeheader}{HTML}{315D7A}
\definecolor{probesubheader}{HTML}{E8F0F5}

\definecolor{proberaegroup}{HTML}{E6DCF2}
\definecolor{probevraerow}{HTML}{F3EEF9}
\definecolor{probevaegroup}{HTML}{F3E2DE}
\definecolor{probevaerow}{HTML}{FAF1EF}
\definecolor{reconlargegroup}{HTML}{DFEAF1}
\definecolor{reconlargerow}{HTML}{F5F8FA}
\definecolor{reconbenchgroup}{HTML}{F2E7D2}
\definecolor{reconbenchrow}{HTML}{FBF7EE}

\definecolor{chartbaseline}{HTML}{8B8397}
\definecolor{chartours}{HTML}{7C5CB2}

\definecolor{findingback}{HTML}{F7F3FC}
\definecolor{findingborder}{HTML}{DCCFEB}
\definecolor{findingaccent}{HTML}{9A7AC7}
\definecolor{findingtag}{HTML}{7253A3}
\definecolor{findingtitle}{HTML}{5C477C}

\newcolumntype{Y}{>{\centering\arraybackslash}X}
\newcolumntype{x}[1]{>{\centering\arraybackslash}p{#1pt}}
\newcolumntype{y}[1]{>{\raggedright\arraybackslash}p{#1pt}}
\newcolumntype{z}[1]{>{\raggedleft\arraybackslash}p{#1pt}}
\newlength\savewidth

\newtcolorbox{promptbox}[1]{
  enhanced,
  breakable,
  colback=black!4,
  colframe=black,
  coltitle=white,
  colbacktitle=black,
  fonttitle=\bfseries\small,
  title=#1,
  boxrule=0.6pt,
  arc=3pt,
  outer arc=3pt,
  top=6pt, bottom=6pt, left=7pt, right=7pt,
  fontupper=\ttfamily\footnotesize\raggedright,
}

\newcommand{\mlogo}{\raisebox{-5pt}{\includegraphics[width=1.5em]{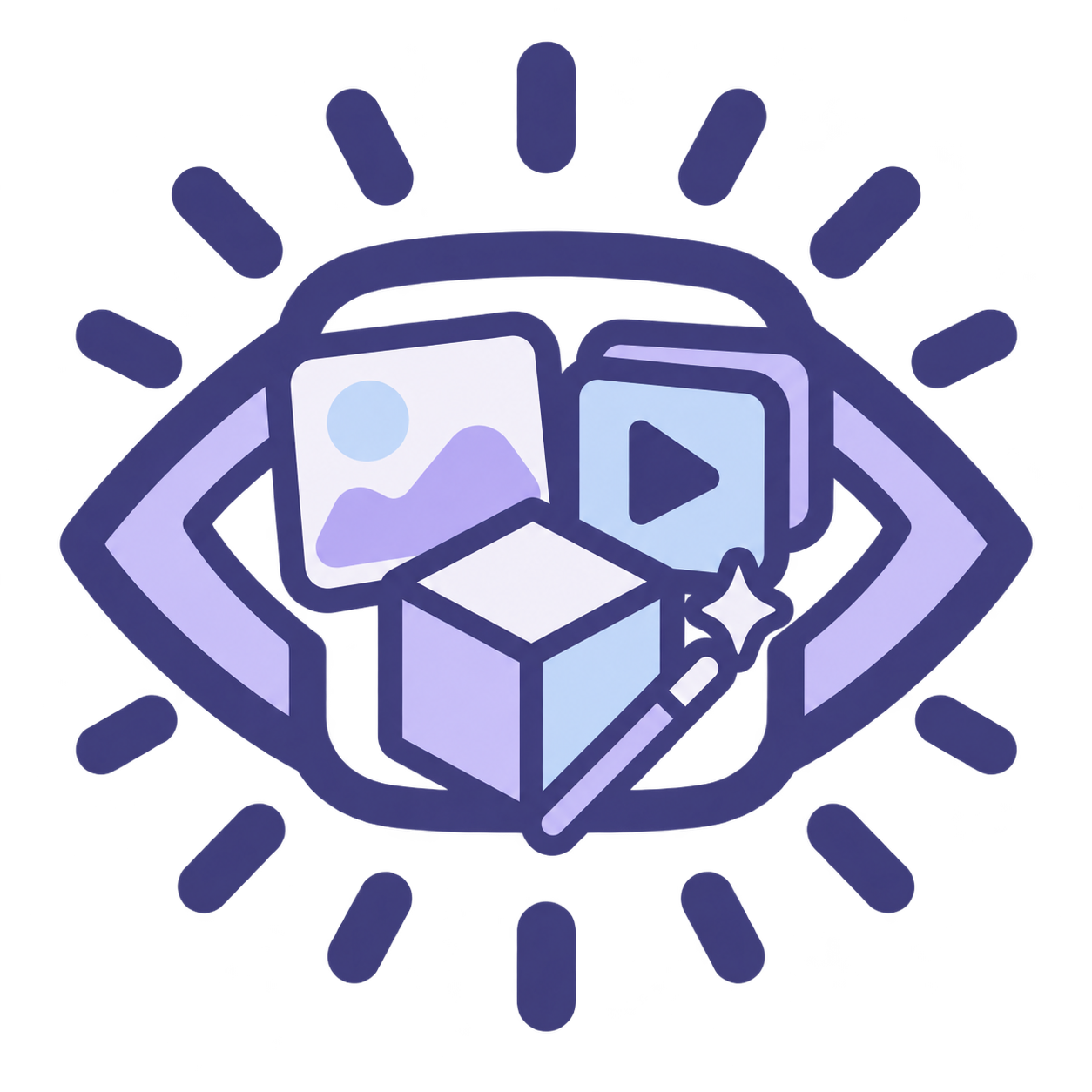}}}

\title{\center{\mlogo\hspace{0.35em}VitronEdit: Universal Visual Editing\\ across Images, Videos, and 3D Scenes}}

\renewcommand\Affilfont{\normalfont\fontsize{11}{15}\selectfont\centering}
\author{%
    \parbox{\textwidth}{%
        \begin{center}
            Jing Wang\textsuperscript{1,\textdagger} \quad
            Yanlin Li\textsuperscript{1,\textdagger} \quad
            Mong-Li Lee\textsuperscript{1} \quad
            Wynne Hsu\textsuperscript{1} \quad
            Shengqiong Wu\textsuperscript{2,*} \quad
            Hao Fei\textsuperscript{2,*} \\
            \textsuperscript{1}National University of Singapore \quad
            \textsuperscript{2}University of Oxford
        \end{center}
    }%
}

\begin{document}

\renewcommand{\abscontent}{%
  \par\vspace{-38pt}
  \noindent\makebox[\linewidth][c]{%
    \large\textbf{Project Page:}~%
    \href{https://vitron-llm.github.io/VitronEdit}{\textcolor{logoindigo}{https://vitron-llm.github.io/VitronEdit}}%
  }%
  \par\vspace{2pt}
  \noindent\begin{minipage}{\linewidth}
    \centering
    \includegraphics[width=0.99\linewidth]{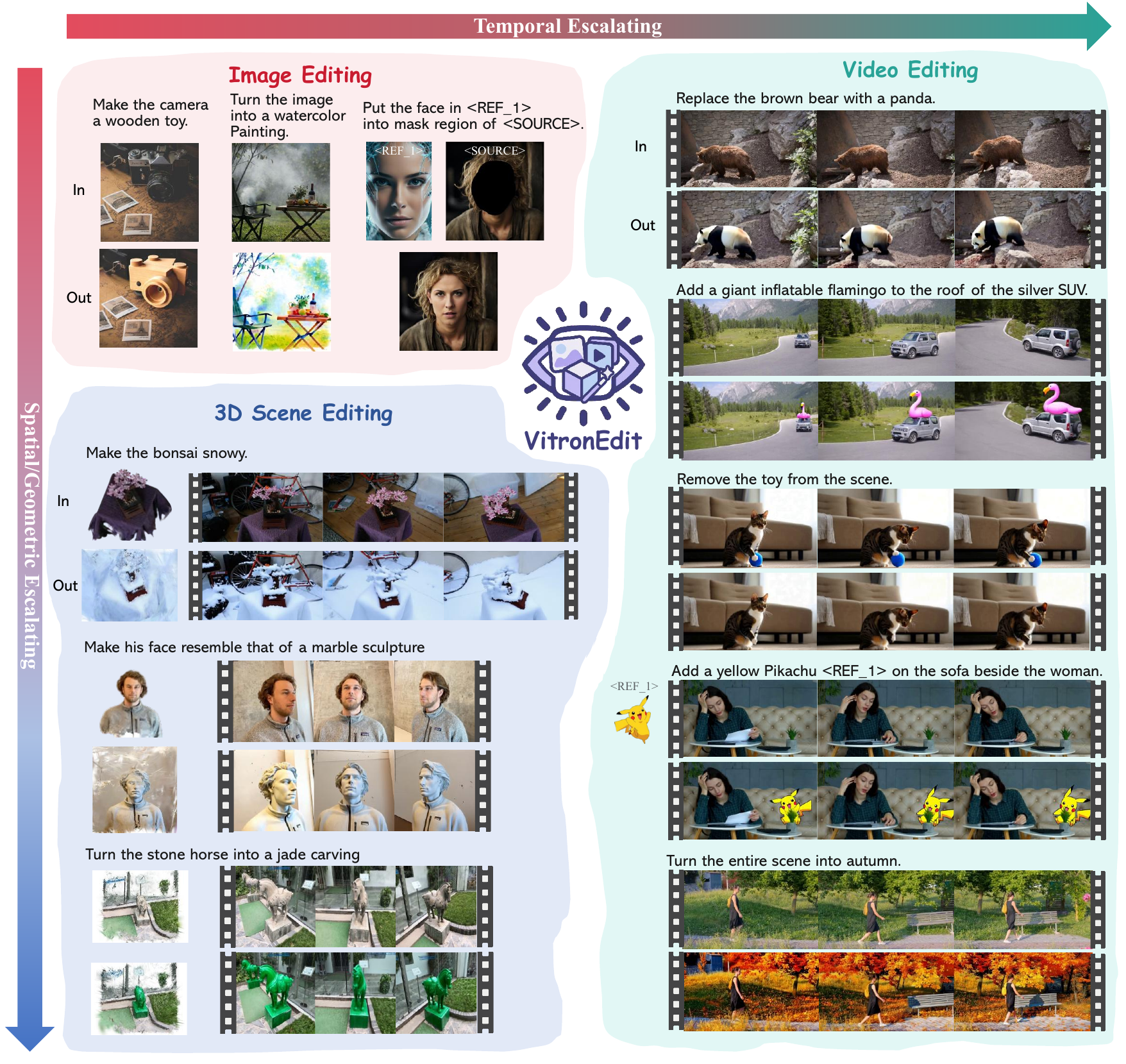}
    \captionof{figure}{Qualitative results of \textbf{VitronEdit} across image, video, and 3D scene editing, demonstrating diverse editing capabilities spanning object manipulation, appearance and style transformations, and increasing temporal and spatial/geometric complexity.}
    \label{fig:case_study_bonsai}
  \end{minipage}%
}

\fancypagestyle{VitronEdit}{%
  \fancyhf{}%
  \fancyfoot[C]{\thepage}%
  \renewcommand{\headrulewidth}{0pt}%
  \renewcommand{\footrulewidth}{0pt}%
}
\fancypagestyle{plain}{%
  \fancyhf{}%
  \fancyfoot[C]{\thepage}%
  \renewcommand{\headrulewidth}{0pt}%
  \renewcommand{\footrulewidth}{0pt}%
}

\fancypagestyle{firststyle}{%
  \fancyhf{}%
  \fancyhead[L]{\small\itshape Preprint, working in progress}%
  \fancyfoot[C]{\thepage}%
  \fancyfoot[L]{%
    \raisebox{18pt}[0pt][0pt]{%
      \begin{minipage}[b]{\textwidth}
        \hrule height0.4pt
        \vspace{4pt}
        \footnotesize\raggedright
        \textsuperscript{\textdagger}Equal contribution.\strut
        Contact author: \href{mailto:jingwang@u.nus.edu}{\texttt{jing\_wang@u.nus.edu}}\strut\\
        \textsuperscript{*}Corresponding authors: Shengqiong Wu (\href{mailto:shengqiongwu@gmail.com}{\texttt{shengqiongwu@gmail.com}})  and Hao Fei (\href{mailto:haofei7419@gmail.com}{\texttt{haofei7419@gmail.com}}).\strut
      \end{minipage}%
    }%
  }%
  \renewcommand{\headrulewidth}{0.4pt}%
  \renewcommand{\footrulewidth}{0pt}%
}

\begin{abstract}
Recent advances in unified visual editing have begun to consolidate image and video editing within a single model, while 3D scene editing remains largely separate and constrained by scarce paired training data.
Given the rich image and video editing supervision now available and the common editing semantics shared across modalities, a natural question is whether this knowledge can be transferred to data-scarce 3D editing.
We present \textbf{VitronEdit}, a universal framework for image, video, and 3D scene editing.
VitronEdit adopts a shared sequence representation and progressive multimodal training to transfer editing knowledge across modalities.
We further introduce \textit{geometry-aware post-training}, which leverages a pretrained geometry reconstruction model to provide depth and camera-pose supervision.
Surprisingly, applying geometric supervision only to 3D samples yields the best overall performance, while image and video editing also improve despite receiving no direct geometric supervision, suggesting that geometric regularization transfers through the shared backbone.
Our analysis further shows that image and video training mainly transfers editing semantics to 3D, whereas 3D supervision primarily improves multi-view consistency.
\end{abstract}
\thispagestyle{firststyle}
\maketitle

\pagestyle{VitronEdit}
\thispagestyle{firststyle}
\clearpage
\thispagestyle{VitronEdit}
\vspace{1ex}
\centerline{\fontsize{15pt}{14pt}\selectfont\textbf{Abstract}}
\vspace{1ex}
{\absfont\noindent\theabstract\par}
\vspace{15pt}

\vspace{-2mm}
\section{Introduction}
\vspace{-2mm}
\label{sec:intro}

Instruction-driven visual editing has rapidly progressed from images to videos and 3D scenes.
Recent works have begun to unify image and video editing within a single model, enabling editing capabilities to be shared across the two modalities~\citep{ju2026editverse,mou2025instructx,xia2025dreamve,wei2026univideo}.
Meanwhile, generalizable 3D editing increasingly builds on pretrained image or video models to propagate edits consistently across multiple viewpoints~\citep{zhao2026tinker,chen2026ViP3DE,huang2025edit360,chen2026omni3dedit}.
These developments suggest that editing knowledge may extend beyond individual visual modalities.
This motivates a broader question:
\emph{can image, video, and 3D editing be learned jointly within a single model, allowing knowledge to be shared across all three domains?}

\begin{figure}[t]
\centering
\includegraphics[width=0.99\linewidth]{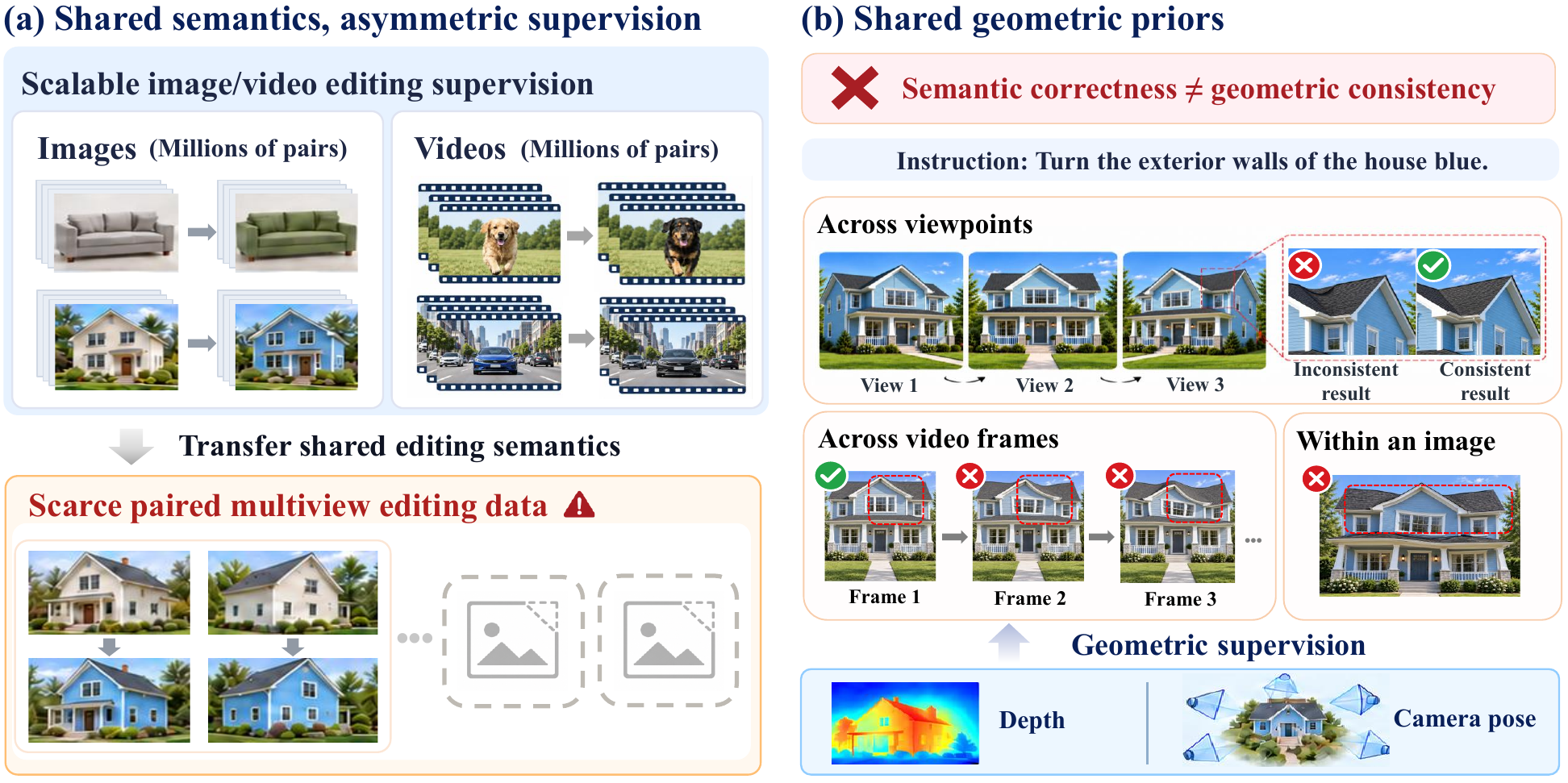}
\vspace{-2mm}
\caption{
Image and video editing data are abundant, while paired multiview data are scarce, and semantically correct edits can still be geometrically inconsistent.
}
\label{fig:motivation}
\vspace{-4mm}
\end{figure}

Two complementary opportunities make such unification promising, as illustrated in Figure~\ref{fig:motivation}.
\textit{First, editing semantics are largely shared across modalities, while editing supervision is highly asymmetric.}
Editing operations such as object addition and removal, appearance modification, and style transfer share common semantics across modalities, providing a basis for transferring editing knowledge~\citep{ju2026editverse,mou2025instructx}.
At the same time, large-scale image and video editing datasets are increasingly available~\citep{brooks2023instructpix2pix,yu2025anyedit,wu2025insvie,bai2026ditto}, whereas high-quality paired multiview editing data remain scarce~\citep{chen2026omni3dedit,wang2026rl3dedit}.
This combination of shared semantics and asymmetric supervision makes cross-modal transfer particularly valuable: a unified model can learn general editing semantics from image and video data, while using comparatively limited 3D data to learn how these edits should remain consistent across viewpoints.

\textit{Second, shared editing semantics do not by themselves ensure geometric consistency.}
Image editing requires geometrically valid spatial structure, while video and multiview editing further require consistency across frames or viewpoints.
Prior work has shown that explicit geometric cues and geometry foundation models can improve structural consistency in image, video, and multiview generation~\citep{wu2024gaussctrl,li2025syncnoise,parihar2025depthaware,du2026videogpa,yin2026vigor,wang2026rl3dedit,ma2026groupediting}.
This suggests that geometric knowledge can serve as a shared structural prior across modalities, complementing semantic transfer within a unified editor.

Based on these insights, we introduce \textbf{VitronEdit}, a universal framework for image, video, and 3D editing. 
VitronEdit represents all three modalities as conditional frame sequences and jointly trains them with a shared backbone and modality-specific tokens, enabling editing knowledge from image and video data to transfer to data-scarce multiview editing.
We further introduce \textit{geometry-aware post-training} with depth and camera-pose supervision provided by a frozen geometry teacher, and investigate how geometric objectives should be applied across the three modalities.
Together, VitronEdit combines cross-modal editing transfer with explicit geometric regularization in a single model.

Our experiments reveal a clear division of roles across modalities. 
Image and video training transfers general editing semantics to multi-view inputs even without 3D supervision, whereas adding limited 3D data mainly strengthens cross-view consistency. 
As a result, VitronEdit outperforms dedicated 3D editing baselines in multi-view consistency while remaining competitive on image and video editing. 
We further find that geometry-aware post-training works best when restricted to 3D samples, improving overall 3D editing performance while also yielding gains on image and video through the shared backbone.

The main contributions of this paper are:
\vspace{-2mm}
\begin{itemize}[leftmargin=2em]
\item We introduce VitronEdit, a unified framework that formulates image, video, and multiview 3D editing as conditional frame-sequence generation within a single shared model.

\item We jointly train VitronEdit across modalities to transfer editing knowledge from images and videos to data-scarce multiview editing, and use geometry-aware post-training that distills depth and camera cues from a frozen geometry teacher to regularize structural consistency.

\item We evaluate VitronEdit on image, video, and 3D editing, showing cross-modal transfer under limited 3D supervision and the effect of geometry-aware post-training under different supervision scopes.
\end{itemize}

\vspace{-1mm}
\section{Related Work}
\vspace{-1mm}
\label{sec:related}

\noindent \textbf{Unified Visual Editing.}
Early attempts at cross-modal visual editing adapt pretrained image diffusion models to structured visual inputs such as panoramas, videos, and 3D scenes through shared inference-time mechanisms~\citep{kim2023collaborative,kwon2024unified}.
While these methods provide a common editing procedure across modalities, they do not learn shared editing capabilities through joint multimodal training.
More recent works unify image and video generation and editing within a single model using shared sequence representations and generative backbones~\citep{ju2026editverse,wei2026univideo,mou2025instructx,chen2026vino,pan2026omniweaving}, enabling editing knowledge to transfer between images and videos.
Unified visual models have also begun to incorporate geometric prediction and novel-view synthesis~\citep{yang2026context}, but instruction-driven 3D scene editing remains largely separate from image--video editing.
VitronEdit bridges this gap by jointly learning image, video, and multiview 3D editing within a single shared model.

\noindent \textbf{3D Scene Editing.}
3D editing commonly transfers powerful 2D editing priors to multiple rendered views and propagates the edits back to an underlying 3D representation.
Early methods iteratively optimize NeRFs or 3D Gaussians using edited views~\citep{haque2023instructnerf2nerf,chen2024gaussianeditor}, with later approaches improving cross-view consistency through joint attention, feature fusion, or explicit geometric constraints~\citep{chen2024dge,lee2025editsplat,wu2024gaussctrl,li2025syncnoise}.
Despite strong editing quality, these approaches typically require per-scene optimization or iterative 2D-3D updates.
Recent methods instead learn generalizable multiview editors or exploit pretrained video priors to generate consistent edited views in a feed-forward manner~\citep{zhao2026tinker,chen2026ViP3DE,huang2025edit360,chen2026omni3dedit,wang2026rl3dedit,zhu2026jointedit3d}.
Related work further shows that treating multiview observations as pseudo-video sequences can combine temporal priors with explicit geometric correspondences~\citep{ma2026groupediting}.
These methods adapt pretrained models into dedicated 3D editors, whereas VitronEdit keeps image and video editing data as joint supervision and retains all three tasks in one model.

\noindent \textbf{Geometry-Aware Visual Generation.}
Generative models trained primarily for visual fidelity may still produce geometrically inconsistent structures, motivating the integration of explicit geometric supervision.
One line of work injects geometric information directly into generative models through depth conditioning, geometric feature alignment, or auxiliary dense prediction~\citep{wu2024gaussctrl,wu2026geometryforcing,bai2025geovideo,rao2026dense}.
A complementary line of work uses pretrained geometry or reconstruction models as teachers, evaluators, or reward models for generative post-training~\citep{go2026textto3d,du2026videogpa,yin2026vigor,an2026vggrpo,wang2026rl3dedit}.
Such approaches demonstrate that geometry foundation models can provide effective supervisory signals beyond conventional reconstruction objectives.
VitronEdit extends this idea to unified visual editing, using a frozen geometry teacher to derive target-conditioned depth and camera supervision applicable to image, video, and multiview editing, and studying its supervision scope.

\section{Method}
\vspace{-1mm}
\label{sec:method}

\subsection{Unified Sequence Representation}
\label{sec:unified-seq}

VitronEdit unifies image, video, and 3D editing within a shared frame-sequence representation. Across modalities, we denote the source and target samples as $x_{\mathrm{src}}^m$ and $x_{\mathrm{tgt}}^m$, where $m \in \{\mathrm{image}, \mathrm{video}, \mathrm{3D}\}$. 
Images are represented as static frame sequences, videos preserve their temporal frame order, and 3D scenes are represented by multi-view sequences ordered according to the sampled camera trajectory. 
This formulation allows all three modalities to share the same backbone and sequence-processing pipeline.
However, the shared sequence representation does not explicitly distinguish static images, temporally evolving videos, and multi-view observations. 
We therefore introduce three learnable modality tokens $\tau^m \in \{\tau_{\text{img}}, \tau_{\text{vid}}, \tau_{\text{3D}}\}$, registered in the vocabulary of a multimodal large language model (MLLM), to encode modality identity.
For each sample, the conditioning information is $c^m = (\ell^m,\tau^m,r^m)$, where $\ell^m$ denotes the editing instruction and $r^m$ is an optional visual reference for reference-based editing tasks. 
The three editing tasks can therefore be expressed under a unified conditional generation formulation: 
\begin{equation} 
p_\theta \left( x_{\mathrm{tgt}}^m \mid x_{\mathrm{src}}^m, c^m \right). 
\label{eq:cond-gen} 
\end{equation}

\begin{figure}[!t]
\centering
\includegraphics[width=0.99\linewidth,trim=185bp 64bp 153bp 55bp,clip]{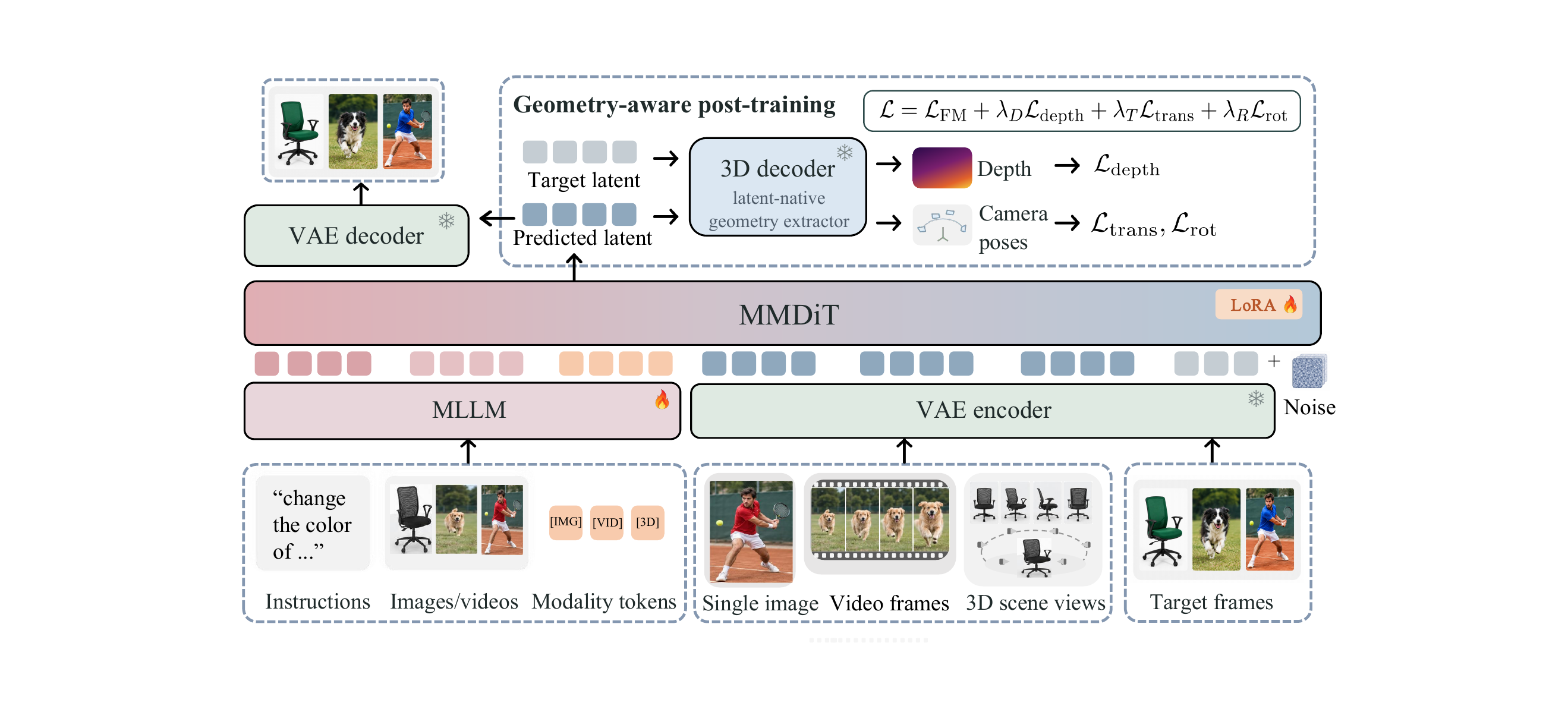}
\vspace{-2mm}
\caption{Overview of VitronEdit. Geometry-aware post-training uses a frozen 3D decoder to estimate depth and camera poses from predicted and target latents.}
\label{fig:arch}
\vspace{-4mm}
\end{figure}

\subsection{Model Architecture}
\label{sec:arch}

As shown in Figure~\ref{fig:arch}, given a source sample $x_{\mathrm{src}}^m$, an editing instruction $\ell^m$, a modality token $\tau^m$, and an optional reference $r^m$, the MLLM encodes the textual instruction together with sampled source and reference frames into semantic conditioning features:
\begin{equation} 
h^m = E_{\mathrm{MLLM}} \left( \ell^m, \tau^m, x_{\mathrm{src}}^m, r^m \right), 
\label{eq:mllm-cond} 
\end{equation} 
where the reference input is omitted when unavailable.
In parallel, a video VAE encodes the complete source and reference sequences into dense visual latents: 
\begin{equation} 
z_{\mathrm{src}}^m = E_{\mathrm{VAE}}(x_{\mathrm{src}}^m), \qquad z_{\mathrm{ref}}^m = E_{\mathrm{VAE}}(r^m), 
\label{eq:visual-cond} 
\end{equation} 
where $z_{\mathrm{ref}}^m$ is omitted for tasks without a visual reference.
The MMDiT serves as the shared generative backbone, jointly incorporating the semantic features (i.e., $h^m $) and dense visual latents to generate the edited target latents.
Edited images and videos are directly decoded by the VAE decoder. 
For 3D editing, the generated latents can either be directly decoded into a 3D Gaussian representation using a 3D decoder or first decoded into multi-view images by the VAE decoder, followed by 3D scene reconstruction.

\subsection{Progressive Cross-Modal Training}
\label{sec:progressive-training}
We optimize all three modalities using the same conditional flow-matching objective~\citep{lipman2023flow}. 
Given a target latent $z_0$, we sample an intermediate latent $z_t$ along the flow path at timestep $t$ and train the model to predict the corresponding target velocity $v_t$: 
\begin{equation}
\mathcal{L}_{\text{FM}} = \mathbb{E}\left[\left\| v_\theta(z_t, t, z^m_{\text{src}}, z^m_{\text{ref}}, h^m) - v_t \right\|_2^2\right],
\end{equation}
where $h^m$ encodes the conditioning information $c^m$ defined in Section~\ref{sec:unified-seq}.
We initialize from a model with existing video-editing capabilities and adopt a progressive training strategy.
We first train on image-editing data, then incorporate video-editing data to restore video-editing capability while retaining image editing, and finally add 3D multi-view editing data. 
At each phase, previously introduced modalities remain in the training mixture, and the corresponding modality tokens are learned jointly.

\subsection{Geometry-Aware Post-Training}
\label{sec:geo}

The unified sequence representation enables knowledge sharing across modalities, but does not explicitly enforce the spatial structure or geometric consistency of the edited outputs. 
We therefore introduce geometry-aware post-training using a frozen latent-native geometry extractor built with the model-stitching strategy of~\citet{go2026textto3d}. 
The extractor predicts depth maps and camera poses directly from both predicted and target latents, allowing the geometry extracted from the target latents to supervise the predicted geometry.

\paragraph{Depth consistency.} 
Inspired by the scale-normalized geometric supervision~\citep{wang2024dust3r,keetha2026mapanything}, we adopt a scale-invariant depth loss. 
We independently normalize the predicted and target depth maps by their sequence-level mean z-depth, denoted by $d_{\mathrm{pred}}$ and $d_{\mathrm{tgt}}$, respectively, and apply a Smooth L1 loss $\rho$:
\begin{equation}
\mathcal{L}_{\mathrm{depth}} = \mathbb{E}_{p} \left[ \rho \left( \frac{D_{\mathrm{pred}}(p)}{d_{\mathrm{pred}}} - \frac{D_{\mathrm{tgt}}(p)}{d_{\mathrm{tgt}}} \right) \right].
\label{eq:depth}
\end{equation}

\paragraph{Camera trajectory consistency.} 
For video and 3D multi-view sequences, we additionally constrain camera trajectories. 
We first align the predicted and target trajectories to their respective first frames and supervise both rotation and translation. 
For rotation, we measure the discrepancy between predicted and target orientations using the scaled chordal distance~\citep{hartley2013rotation}:
\begin{equation}
\mathcal{L}_{\text{rot}} = \mathbb{E}_{i \geq 1}\left[\frac{\left\| \bar{R}^{\text{pred}}_i - \bar{R}^{\text{tgt}}_i \right\|_F}{\sqrt{2}}\right].
\label{eq:rot}
\end{equation}
For translation, we supervise both adjacent-frame displacements and first-frame-relative positions: displacement errors can accumulate into
trajectory drift, whereas position supervision alone provides weaker local constraints~\citep{sturm2012benchmark}. 
Let $\bar{t}_i$ denote the first-frame-aligned camera translation and $d$ the sequence-level mean depth.
We define $\tilde{t}_i = \bar{t}_i / d$ and
$\Delta\tilde{t}_i = \tilde{t}_i - \tilde{t}_{i-1}$, and optimize
\begin{equation}
\mathcal L_{\mathrm{trans}} = \mathbb E_i \left[ \lambda_{\mathrm{rel}} \rho \left( \Delta \tilde t_i^{\mathrm{pred}} - \Delta \tilde t_i^{\mathrm{tgt}} \right) + \lambda_{\mathrm{abs}} \rho \left( \tilde t_i^{\mathrm{pred}} - \tilde t_i^{\mathrm{tgt}} \right) \right],
\label{eq:trans}
\end{equation}
where the first term constrains local motion and the second anchors the global trajectory. Coordinate alignment and scale normalization are detailed in Appendix~\ref{app:camera_loss}.
The overall objective combines the flow-matching and geometric losses:
\begin{equation}
\mathcal{L} = \mathcal{L}_{\mathrm{FM}} + \lambda_D\mathcal{L}_{\mathrm{depth}} + \lambda_T\mathcal{L}_{\mathrm{trans}} + \lambda_R\mathcal{L}_{\mathrm{rot}}.
\label{eq:total}
\end{equation}
We consider depth supervision for image editing, and both depth and camera supervision for video and 3D editing. 
We then study how these geometric objectives should be applied across modalities. 
In our final configuration, both depth and camera supervision are applied only to 3D samples. Appendix~\ref{app:geo_loss} provides additional normalization details and hyperparameters.

\section{Experimental Settings}
\label{sec:exp}

\vspace{-1mm}
\paragraph{Datasets.}
We curate the training data from existing image, video, and 3D datasets~\citep{yu2025anyedit,kuprashevich2026nohumansrequired,song2026insertanything,he2026openve,lin2026kiwiedit,ye2026nano3d,ling2024dl3dv}.
Since large-scale 3D scene editing data are not readily available, we additionally construct a multiview 3D scene editing dataset with substantial viewpoint variation, comprising 8K samples.
The construction pipeline is provided in Appendix~\ref{app:multiview_data}.
In total, our training set contains approximately 2.53M samples.
The detailed data composition is summarized in Table~\ref{tab:training_data} in Appendix~\ref{app:training}.

\vspace{-2mm}
\paragraph{Baselines and evaluation metrics.}
For image editing, we compare against instruction-based methods ICEdit~\citep{zhang2025icedit}, VAREdit~\citep{mao2026varedit}, and ChordEdit~\citep{lu2026chordedit}, as well as ACE++~\citep{mao2025aceplus} and FLUX.1 Kontext~\citep{labs2025flux}, which additionally support reference-conditioned editing.
For video editing, we include the instruction-based baselines VideoCoF~\citep{yang2026videocof} and Omni-Video 2~\citep{yang2026omni}, together with VACE~\citep{jiang2025vace}, UniVideo~\citep{wei2026univideo}, VINO~\citep{chen2026vino}, Kiwi-Edit~\citep{lin2026kiwiedit}, and OmniWeaving~\citep{pan2026omniweaving}, which support both instruction-based and reference-conditioned editing.
For 3D editing, we compare with GaussianEditor~\citep{chen2024gaussianeditor}, DGE~\citep{chen2024dge}, and EditSplat~\citep{lee2025editsplat}.
We evaluate image editing in terms of structural preservation and semantic alignment with the editing instruction.
Video editing is additionally assessed for temporal consistency and visual quality, while 3D editing focuses on instruction alignment and multi-view consistency.
Detailed evaluation protocols are provided in Appendix~\ref{app:eval-protocols}.

\vspace{-2mm}
\paragraph{Training details.}
We initialize VitronEdit with OmniWeaving~\citep{pan2026omniweaving}, which adopts HunyuanVideo-1.5~\citep{wu2025hunyuanvideo} as the generative backbone and Qwen2.5-VL~\citep{bai2025qwen25vl} as the MLLM.
We apply LoRA~\citep{hu2022lora} to the DiT backbone with a rank of 32 and a scaling factor $\alpha=32$, and extend the MLLM vocabulary with modality-specific tokens.
Stage 1 follows a progressive multimodal training schedule: we first train on image-editing data for 10K steps, then introduce video-editing data for an additional 7K steps, and finally incorporate 3D multiview editing data for 3K steps.
Starting from the Stage-1 checkpoint, Stage 2 performs geometry-aware post-training for 500 steps with a frozen latent-native geometry extractor constructed following the model-stitching strategy of~\citet{go2026textto3d}.
Further training details are provided in Appendix~\ref{app:training}.

\vspace{-1mm}
\begin{table}[!t]
\centering
\captionof{table}{Quantitative comparison on \textit{\textbf{Image Editing}}. Struct. Dist.: Structure Distance. Ref.: Reference. Best in \textbf{bold}, second-best \underline{underlined}.}
\label{tab:image-editing}
\fontsize{9}{9}\selectfont
\setlength{\tabcolsep}{2.7mm}
\renewcommand{\arraystretch}{1.1}
\begin{tabular}{@{}lcccccccc@{}}
\toprule
\multirow[c]{3}{*}{\textbf{Method}} &
\multicolumn{5}{c}{\textbf{Consistency}} &
\multicolumn{3}{c}{\textbf{CLIP Semantics}} \\
\cmidrule(lr){2-6} \cmidrule(lr){7-9}
& \textbf{Struct. Dist.} $\downarrow$
& \textbf{PSNR} $\uparrow$
& \textbf{MSE} $\downarrow$
& \textbf{LPIPS} $\downarrow$
& \textbf{SSIM} $\uparrow$
& \textbf{Whole} $\uparrow$
& \textbf{Edited} $\uparrow$
& \textbf{Ref.} $\uparrow$ \\
& $\times 10^2$
& \phantom{$\times 10^2$}
& $\times 10^3$
& $\times 10^2$
& $\times 10^2$
& $\times 10^2$
& $\times 10^2$
& $\times 10^2$ \\
\midrule
\rowcolor{gray!15}
\multicolumn{9}{c}{\textbf{\textit{Instruction-based editing}}} \\
\midrule
ICEdit     & \underline{3.54}  & \textbf{27.88} & \textbf{6.01}  & \textbf{5.76}  & \textbf{89.39} & 24.24 & 21.94 & /  
\\
VAREdit    & 5.75  & 23.52 & 11.02 & \underline{8.19}  & 81.75 & \underline{25.47} & \textbf{22.74} & /  
\\
ChordEdit  & \textbf{2.88}  & 22.60 & \underline{8.04}  & 12.35 & 76.52 & 24.59 & 22.06 & /  \\

VitronEdit (Ours)       &  8.44 & \underline{24.20} & 17.19 & 10.07 & \underline{85.84} & \textbf{25.70} & \underline{22.45} & /  
\\
\midrule
\rowcolor{gray!15}
\multicolumn{9}{c}{\textbf{\textit{Instruction-based editing \& Reference-based editing}}} \\
\midrule
ACE++           & \underline{10.98} & \underline{27.75} & \underline{9.20}  & \underline{5.99}  & \underline{86.43} & 21.07 & 18.79 & \underline{67.96} \\
FLUX.1 Kontext  & \textbf{9.26}  & \textbf{36.49} & \textbf{0.32}  & \textbf{1.44}  & \textbf{97.07} & \textbf{24.27} & \textbf{21.92} & \textbf{78.68} \\

\textbf{VitronEdit (Ours)}            & 12.39 & 21.95 & 29.12 & 14.86 & 81.91 & \underline{23.55} & \underline{21.10} & 64.40 \\
\bottomrule
\end{tabular}
\vspace{-2mm}
\end{table}

\begin{figure}[!t]
\centering
\includegraphics[width=0.99\linewidth]{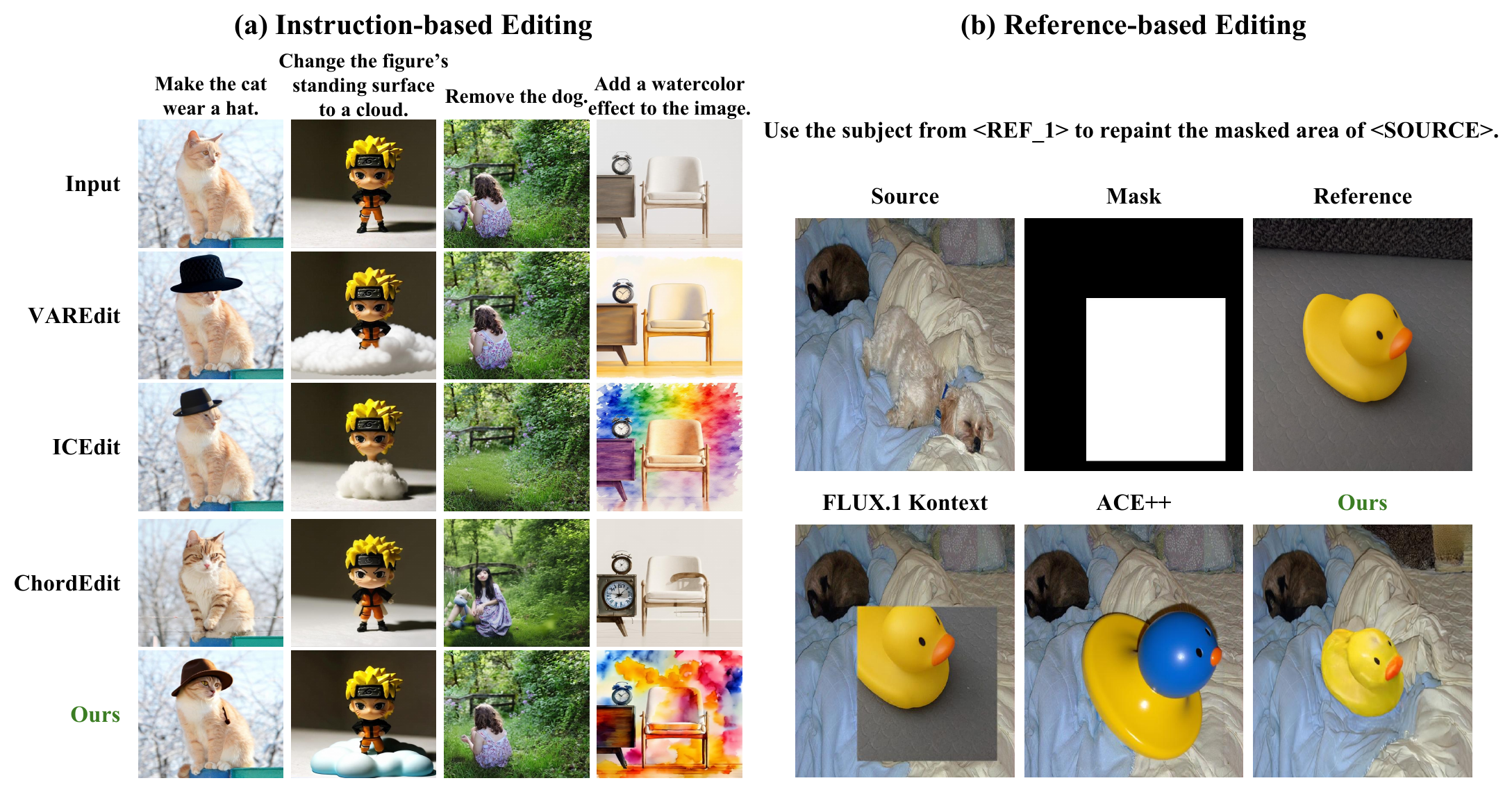}
\vspace{-2mm}
\captionof{figure}{
Comparison with image-editing baselines: (a) instruction-based, (b) reference-based.
}
\label{fig:case_study_image}
\vspace{-2mm}
\par
\end{figure}

\section{Results and Discussion}
\label{sec:main-results}

In our experiments, we want to address four questions as follows:
\vspace{-2mm}
\begin{itemize}[leftmargin=1.5em]
    \item \textbf{RQ-1}: How does VitronEdit perform across image, video, and 3D editing?
    \item \textbf{RQ-2}: How do image/video training and subsequent 3D training contribute to 3D editing?
    \item \textbf{RQ-3}: How do the type and scope of geometric supervision affect editing performance across modalities?
    \item \textbf{RQ-4}: How does sequence construction affect detail preservation in image and 3D editing?
\end{itemize}

\subsection{Editing Performance across Modalities (RQ-1)}

\paragraph{Image Editing.}
First, we examine whether VitronEdit retains strong editing capabilities when jointly trained across image, video, and 3D modalities, as shown in Table~\ref{tab:image-editing}.
For instruction-based editing, VitronEdit achieves the best whole-image CLIP score and the second-best edited-region CLIP score, indicating strong alignment with the requested edits.
It also ranks second in PSNR and SSIM, although its larger structure distance and pixel-level errors suggest less reliable preservation of fine-grained source content than specialized image editors.
The qualitative examples in Figure~\ref{fig:case_study_image}(a) further show that VitronEdit can perform diverse object, appearance, and style edits while following the instruction.
For reference-based editing, VitronEdit remains competitive in semantic alignment but trails specialized baselines in both source-content preservation and reference similarity.
Figure~\ref{fig:case_study_image}(b) illustrates a complementary qualitative behavior: VitronEdit adapts the reference subject to the source scene, whereas FLUX.1 Kontext also transfers part of the reference background into the edited region.
Overall, VitronEdit preserves strong instruction-following ability within the unified framework, while precise content preservation and reference fidelity remain the main gaps relative to dedicated image-editing models.

\begin{table}[!t]
\centering
\caption{Quantitative comparison on \textit{\textbf{Video Editing}}. Best in \textbf{bold}, second-best \underline{underlined}.}
\label{tab:video-editing}
\fontsize{9}{9}\selectfont
\setlength{\tabcolsep}{1.5mm}
\renewcommand{\arraystretch}{1.1}
\makebox[\textwidth][c]{%
\begin{tabular}{@{}lccccccccc@{}}
\toprule
\multirow[c]{3}{*}{\textbf{Method}} &
\multicolumn{3}{c}{\textbf{Consistency}} &
\multicolumn{3}{c}{\textbf{CLIP Semantics}} &
\multicolumn{3}{c}{\textbf{Temporal Quality}} \\
\cmidrule(lr){2-4} \cmidrule(lr){5-7} \cmidrule(lr){8-10}
& \textbf{PSNR} $\uparrow$
& \textbf{LPIPS} $\downarrow$
& \textbf{SSIM} $\uparrow$
& \textbf{Whole} $\uparrow$
& \textbf{Edited} $\uparrow$
& \textbf{Ref.} $\uparrow$
& \textbf{MFS} $\uparrow$
& \textbf{MFS-Edit} $\uparrow$
& \textbf{DOVER} $\uparrow$ \\
& \phantom{$\times 10^2$}
& $\times 10^2$
& $\times 10^2$
& $\times 10^2$
& $\times 10^2$
& $\times 10^2$
& $\times 10^2$
& $\times 10^2$
& $\times 10^2$ \\
\midrule
\rowcolor{gray!15}
\multicolumn{10}{c}{\textbf{\textit{Instruction-based editing}}} \\
\midrule
VideoCoF      & \textbf{25.64} & \textbf{10.23} & \textbf{81.23} & \underline{26.97} & \textbf{21.44} & /  & \textbf{85.42} & \textbf{83.69} & \textbf{58.46} \\
OmniVideo 2   & 11.28 & 49.92 & 39.39 & 23.02 & 18.83 & /  & 35.61 & 33.28 & 28.85 \\
\textbf{Ours} & \underline{17.33} & \underline{33.87} & \underline{61.21} & \textbf{27.14} & \underline{20.50} & /  & \underline{76.70} & \underline{74.56} & \underline{48.41} \\

\midrule
\rowcolor{gray!15}
\multicolumn{10}{c}{\textbf{\textit{Instruction-based editing \& Reference-based editing}}} \\
\midrule
VACE         & 17.42 & \underline{28.16} & 59.65 & 26.54 & 20.49 & \textbf{66.95} & \textbf{82.02} & \textbf{80.67} & \underline{56.44} \\
UniVideo     & 15.56 & 33.69 & 48.59 & \textbf{27.44} & \underline{21.17} & 61.88 & 60.94 & 55.78 & 48.01 \\
VINO         & 14.59 & 30.49 & 47.37 & \underline{26.99} & \textbf{21.47} & \underline{65.99} & 72.99 & 69.96 & \textbf{61.44}   \\
Kiwi-Edit    & \underline{18.49} & \textbf{22.04} & 57.90 & 26.32 & 21.13 & 59.28 & 78.30 & 75.88 & 54.90 \\
OmniWeaving  & \textbf{19.57} & 30.07 & \textbf{65.24} & 26.50 & 20.16 & 55.82 & \underline{80.67} & \underline{78.91} & 44.66 \\

\textbf{VitronEdit (Ours)}         & 17.32 & 33.78 & \underline{61.25} & 26.95 & 20.54 & 60.60 & 76.80 & 74.69 & 48.56 \\

\bottomrule
\end{tabular}
}
\end{table}

\begin{figure}[!t]
\centering
\includegraphics[width=0.99\linewidth]{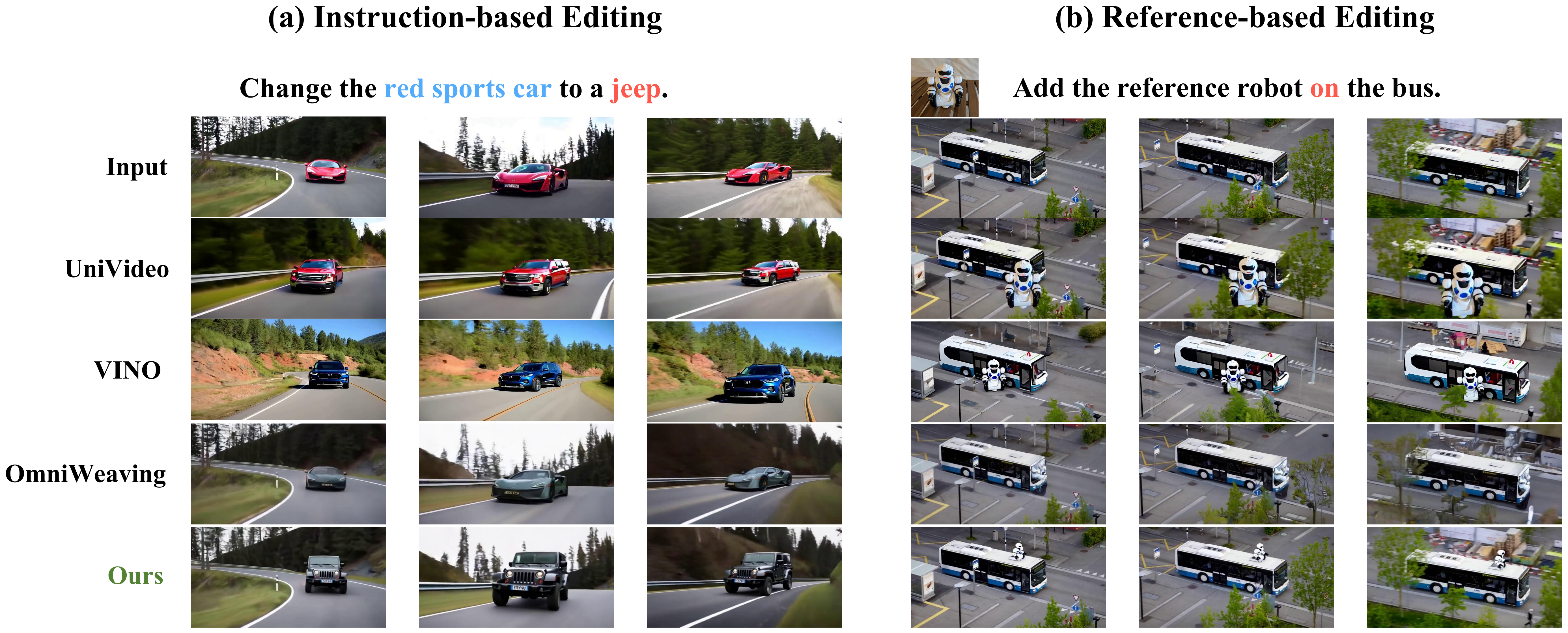}
\vspace{-2mm}
\caption{
Comparison with unified-model baselines (UniVideo, VINO, OmniWeaving): (a) instruction-based, (b) reference-based.
}
\label{fig:case_study_video}
\end{figure}

\vspace{-3mm}
\paragraph{Video Editing.}
We next examine video editing capability.
Table~\ref{tab:video-editing} shows that VitronEdit retains strong semantic editing capability for videos.
For instruction-based editing, it achieves the best whole-video CLIP score and remains competitive in edited-region alignment.
However, VideoCoF performs substantially better on content-preservation and temporal-quality metrics, revealing a gap in preserving the source sequence.
For reference-based editing, VitronEdit remains competitive in semantic alignment but does not lead in reference similarity, content preservation, or temporal fidelity.
Compared with its initialization model, OmniWeaving, VitronEdit improves whole-frame, edited-region, and reference CLIP scores, as well as DOVER, while showing lower PSNR, SSIM, and motion-consistency scores.
Figure~\ref{fig:case_study_video} further illustrates these strengths: VitronEdit consistently realizes the sports-car-to-jeep transformation across frames and preserves the requested ``robot on the bus'' spatial relation, while competing methods often under-edit the object category or misplace the reference subject.

\begin{figure}[!t]
\centering
\includegraphics[width=0.99\linewidth]{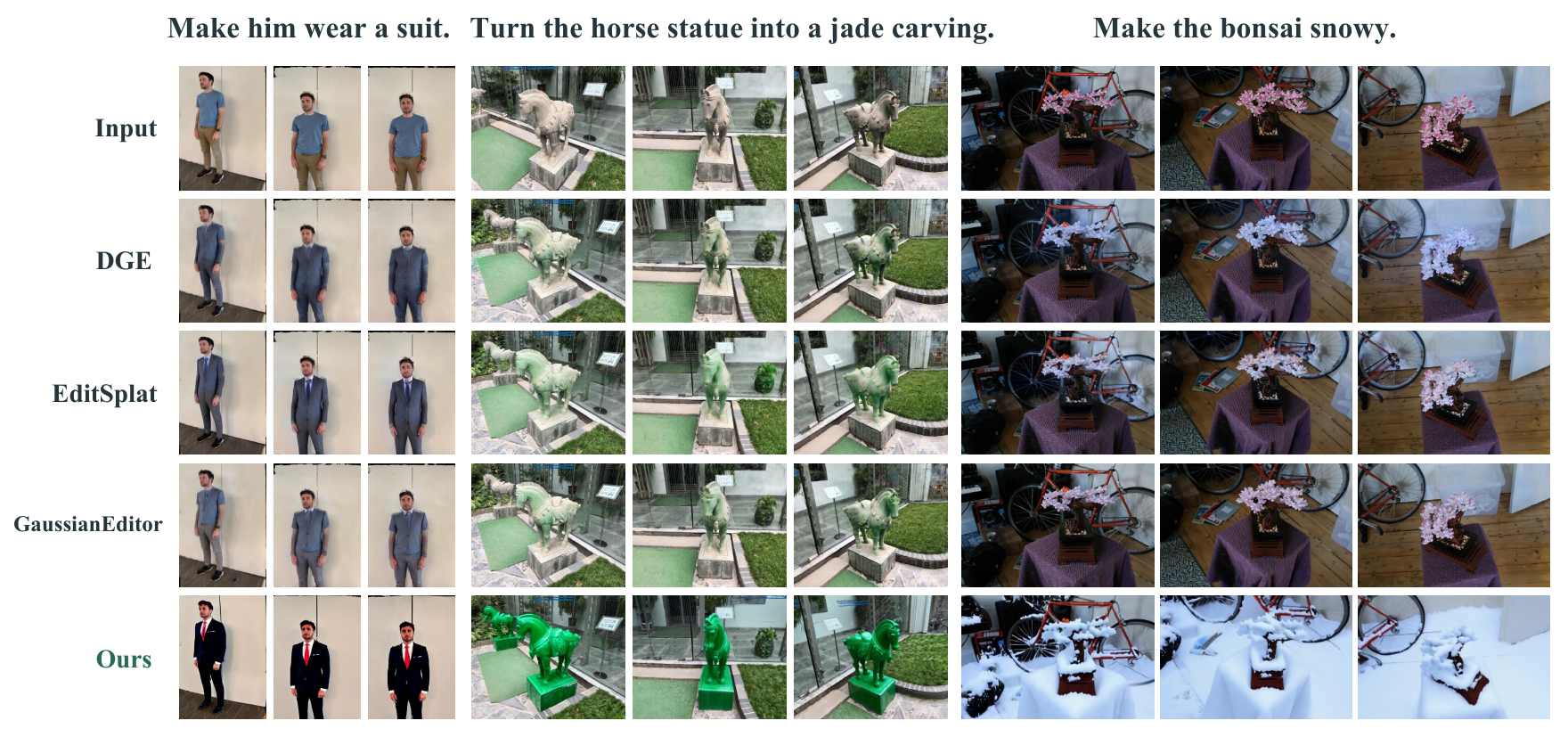}
\vspace{-4mm}
\caption{
Qualitative comparison against all 3D-editing baselines.
}
\label{fig:case_study_3d}
\vspace{-2mm}
\end{figure}

\vspace{-1mm}
\paragraph{3D Editing.}
We evaluate whether VitronEdit can effectively transfer editing capability to 3D while preserving cross-view consistency.
As demonstrated in Table~\ref{tab:3d-editing}, VitronEdit shows the clearest advantage in multi-view consistency compared with baselines.
VitronEdit achieves the best 3D consistency score and the lowest photometric reprojection error, while remaining close to EditSplat in semantic alignment, ranking second on both CLIPDir and CLIPSim.
This indicates that the unified model preserves competitive editing semantics while producing substantially more consistent results across viewpoints.
Figure~\ref{fig:case_study_3d} further illustrates this behavior.
VitronEdit applies clothing, material, and appearance changes consistently across views, as shown by the suit, jade-carving, and snowy-bonsai examples.
The snowy example also exposes a limitation: surrounding regions are modified more broadly than requested, showing that cross-view coherence does not necessarily imply precise editing locality.
Its main advantage therefore lies in maintaining strong semantic editing while substantially improving multi-view consistency.

\begin{table}[!t]
\centering
\caption{Quantitative comparison on \textit{\textbf{3D Editing}}. Best in \textbf{bold}, second-best \underline{underlined}.}
\label{tab:3d-editing}
\fontsize{9}{9}\selectfont
\setlength{\tabcolsep}{3.4mm}
\renewcommand{\arraystretch}{1.1}
\begin{tabular}{@{}lcccc@{}}
\toprule
\textbf{Method}
& \textbf{CLIPDir} $\uparrow$ $\times 10^2$
& \textbf{CLIPSim} $\uparrow$ $\times 10^2$
& \textbf{3D Consistency} $\uparrow$ $\times 10^2$ 
& \textbf{Ph-Loss} $\downarrow$ $\times 10^2$ \\ 
\midrule
GaussianEditor &  9.43 & 26.84 & 64.14 & 20.93 \\
DGE            & 11.27 & 28.12 & 64.79 & 19.86 \\
EditSplat      & \textbf{13.76} & \textbf{28.60} & \underline{66.30} & \underline{19.69} \\

\textbf{VitronEdit (Ours)}  & \underline{13.15} & \underline{28.43} &  \textbf{74.34} & \textbf{18.88} \\
\bottomrule
\end{tabular}
\vspace{-2mm}
\end{table}

\vspace{-2mm}
\subsection{Cross-Modal Transfer to 3D Editing (RQ-2)}
\label{sec:transfer}
\vspace{-1mm}

\begin{figure}[!htbp]
\centering
\includegraphics[width=0.85\linewidth]{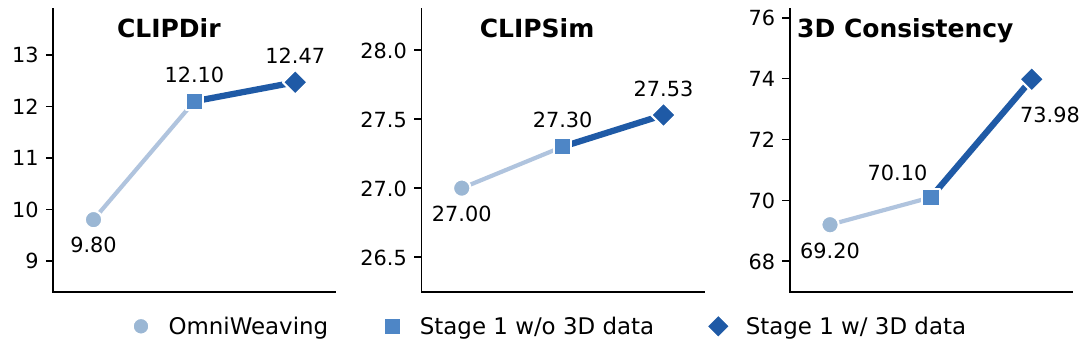}
\vspace{-2mm}
\caption{
Cross-modal transfer to 3D editing. Stage-1 training on image and video data substantially improves editing semantics on multi-view inputs, while incorporating 3D data yields further gains primarily in cross-view consistency.
}
\label{fig:transfer}
\end{figure}

To disentangle cross-modal semantic transfer from 3D-specific consistency learning, we compare the initialization model with Stage-1 training without and with 3D editing data.
As shown in Figure~\ref{fig:transfer}, training only on image and video data substantially improves semantic editing on multiview inputs: CLIPDir increases from 9.80 to 12.10 and CLIPSim from 27.00 to 27.30, while 3D Consistency changes only marginally from 69.20 to 70.10.
This indicates that editing knowledge learned from images and videos transfers effectively to multiview editing even without paired 3D supervision.
Adding 3D editing data produces only modest further gains in CLIPDir and CLIPSim, but raises 3D Consistency substantially from 70.10 to 73.98.
The contrasting gains suggest that image and video supervision primarily transfers editing semantics, whereas limited 3D supervision mainly teaches how those edits should be realized consistently across viewpoints.

\vspace{-1mm}
\subsection{Effect and Scope of Geometric Supervision (RQ-3)}
\label{sec:geo-ablation}
\vspace{-2mm}

We next examine which geometric signals are useful and, more importantly, where they should be applied.
Table~\ref{tab:geo-overall} compares no geometric supervision (\textit{w/o Geo}), depth-only supervision (\textit{w/o Camera}), camera-only supervision (\textit{w/o Depth}), full supervision across all applicable modalities (\textit{Geo-All}), and full supervision restricted to 3D samples (\textit{Geo-3D}).

When geometric supervision is applied broadly, neither depth nor camera cues yield consistent gains across modalities.
Depth-only and camera-only training improve several video metrics, but both reduce CLIPDir and 3D Consistency; applying both signals to all modalities (\textit{Geo-All}) further degrades all reported 3D metrics and provides little benefit to image or video editing.
These results indicate that simply increasing geometric supervision does not necessarily improve the unified editor.

The supervision scope is more consequential.
Restricting both depth and camera losses to 3D samples (\textit{Geo-3D}) improves every reported metric over \textit{w/o Geo} and consistently outperforms \textit{Geo-All}.
Notably, image and video performance also improves despite receiving no direct geometric supervision.
This suggests that geometry learned from 3D data acts as a transferable regularizer through the shared backbone, rather than requiring explicit supervision for every modality.

We attribute this behavior to the alignment between geometric supervision and the underlying data structure.
For multiview 3D scenes, depth and camera poses directly constrain cross-view geometry, whereas the same signals are less tightly aligned with image and video editing, particularly for videos containing object motion.
We therefore adopt \textit{Geo-3D} as the default configuration.

\begin{table}[!t]
\centering
\caption{
Ablation of geometry-aware post-training. Values in parentheses show changes from Ours (Geo-3D), with $\uparrow$ and $\downarrow$ directions considered. \textcolor{red}{Red} indicates degradation after ablation, while \textcolor{gray!130}{gray} indicates improvement. 3D Consist. denotes 3D Consistency.
}
\label{tab:geo-overall}
\fontsize{9}{10}\selectfont
\setlength{\tabcolsep}{1mm}
\renewcommand{\arraystretch}{1.1}
\resizebox{\textwidth}{!}{%
\begin{tabular}{@{}lcccccccc@{}}
\toprule
\multirow[c]{2}{*}{\textbf{Method}} &
\multicolumn{2}{c}{\textbf{Image Editing}} &
\multicolumn{3}{c}{\textbf{Video Editing}} &
\multicolumn{3}{c}{\textbf{3D Editing}} \\
\cmidrule(lr){2-3} \cmidrule(lr){4-6} \cmidrule(lr){7-9}
& \textbf{LPIPS} $\downarrow$
& \textbf{Whole} $\uparrow$
& \textbf{LPIPS} $\downarrow$
& \textbf{Whole} $\uparrow$
& \textbf{MFS} $\uparrow$
& \textbf{CLIPDir} $\uparrow$
& \textbf{CLIPSim} $\uparrow$
& \textbf{3D Consist.} $\uparrow$ \\
\midrule

-\ w/o \textbf{Geo}
& 14.88 {\scriptsize\textcolor{red}{(-0.02)}}
& 23.48 {\scriptsize\textcolor{red}{(-0.07)}}
& 34.20 {\scriptsize\textcolor{red}{(-0.42)}}
& 26.90 {\scriptsize\textcolor{red}{(-0.05)}}
& 76.05 {\scriptsize\textcolor{red}{(-0.75)}}
& 12.47 {\scriptsize\textcolor{red}{(-0.68)}}
& 27.53 {\scriptsize\textcolor{red}{(-0.90)}}
& 73.98 {\scriptsize\textcolor{red}{(-0.36)}} \\

-\ w/o \textbf{Camera}
& 15.02 {\scriptsize\textcolor{red}{(-0.16)}}
& 23.48 {\scriptsize\textcolor{red}{(-0.07)}}
& 34.01 {\scriptsize\textcolor{red}{(-0.23)}}
& 26.92 {\scriptsize\textcolor{red}{(-0.03)}}
& 76.38 {\scriptsize\textcolor{red}{(-0.42)}}
& 11.76 {\scriptsize\textcolor{red}{(-1.39)}}
& 27.49 {\scriptsize\textcolor{red}{(-0.94)}}
& 72.17 {\scriptsize\textcolor{red}{(-2.17)}} \\

-\ w/o \textbf{Depth}
& 15.19 {\scriptsize\textcolor{red}{(-0.33)}}
& 23.57 {\scriptsize\textcolor{gray!130}{(+0.02)}}
& 33.72 {\scriptsize\textcolor{gray!130}{(+0.06)}}
& 26.92 {\scriptsize\textcolor{red}{(-0.03)}}
& 77.05 {\scriptsize\textcolor{gray!130}{(+0.25)}}
& 12.10 {\scriptsize\textcolor{red}{(-1.05)}}
& 27.54 {\scriptsize\textcolor{red}{(-0.89)}}
& 72.45 {\scriptsize\textcolor{red}{(-1.89)}} \\

\midrule

\textbf{Geo-All}
& 14.96 {\scriptsize\textcolor{red}{(-0.10)}}
& 23.50 {\scriptsize\textcolor{red}{(-0.05)}}
& 34.35 {\scriptsize\textcolor{red}{(-0.57)}}
& 26.89 {\scriptsize\textcolor{red}{(-0.06)}}
& 76.07 {\scriptsize\textcolor{red}{(-0.73)}}
& 11.62 {\scriptsize\textcolor{red}{(-1.53)}}
& 27.33 {\scriptsize\textcolor{red}{(-1.10)}}
& 72.82 {\scriptsize\textcolor{red}{(-1.52)}} \\

\textbf{Ours (Geo-3D)}
& 14.86
& 23.55
& 33.78
& 26.95
& 76.80
& 13.15
& 28.43
& 74.34 \\

\bottomrule
\end{tabular}
}%
\vspace{-2mm}
\end{table}

\begin{figure}[!t]
\centering
\begin{minipage}[t]{0.50\linewidth}
\centering
\includegraphics[height=163pt]{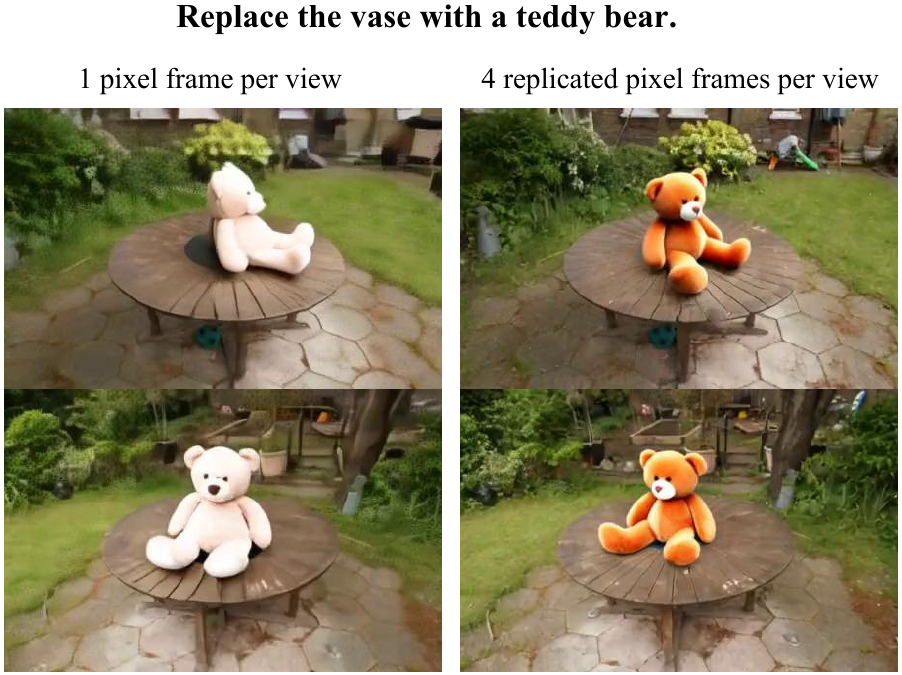}
\vspace{-2mm}
\caption{
Four-frame replication per view (right) preserves sharper, more saturated detail than a single frame per view (left).
}
\label{fig:sequence_ablation_3d}
\end{minipage}
\hfill
\begin{minipage}[t]{0.46\linewidth}
\centering
\includegraphics[height=170pt]{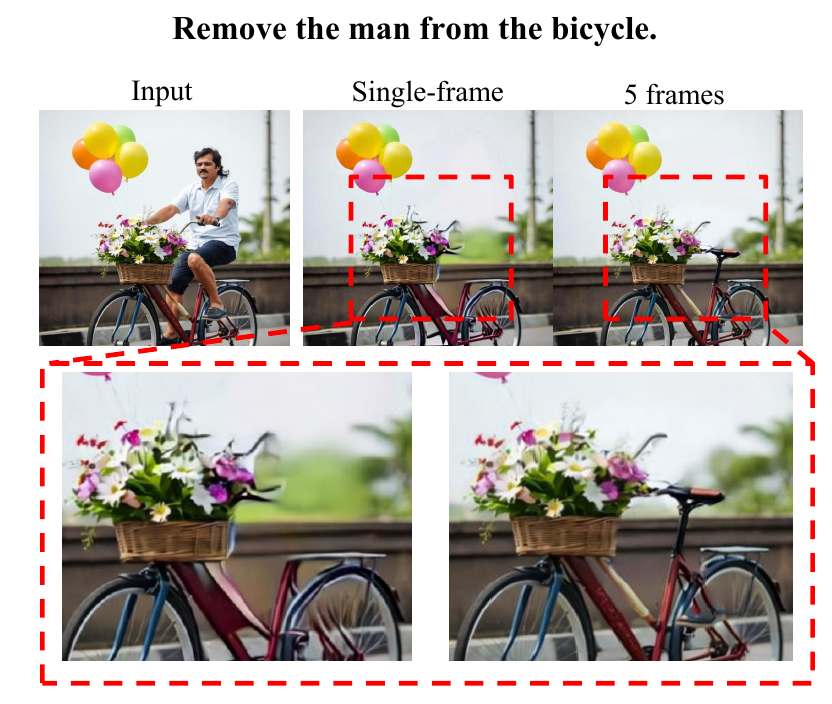}
\vspace{-2mm}
\caption{
Five-frame static replication (right) recovers finer structural detail than single-frame input (middle) for image editing.
}
\label{fig:sequence_ablation_img}
\end{minipage}
\vspace{-2mm}
\end{figure}

\vspace{-2mm}
\subsection{Modality-Specific Sequence Design (RQ-4)}
\label{sec:impl-ablation}
\vspace{-1mm}

Although VitronEdit represents all modalities as frame sequences, directly using the same sequence construction is suboptimal because the pretrained video backbone assumes temporally dense inputs.
We therefore adapt the sequence representation to the characteristics of 3D views and static images.

\noindent \textbf{3D: Replicating Each View.}
We observed noticeable detail loss when directly encoding multiview 3D sequences with the pretrained temporal VAE.
Unlike ordinary videos, adjacent 3D views can exhibit large viewpoint changes, making temporal compression prone to discarding high-frequency details.
Reconstructing the original multiview sequences with the HunyuanVideo-1.5 VAE~\citep{wu2025hunyuanvideo} yields only 26.02 dB PSNR and preserves 41.9\% of the high-frequency content.
To better align multiview inputs with the VAE's temporal compression ratio, we keep the first view once and replicate each subsequent view four times.
This increases reconstruction PSNR to 30.24 dB and high-frequency retention to 66.5\%, with visibly sharper details in Figure~\ref{fig:sequence_ablation_3d}.
We therefore use this representation for 3D training.

\noindent \textbf{Image: Static Sequence Replication.}
A single-frame representation, in contrast, underutilizes a backbone pretrained for multi-frame generation.
We therefore replicate each image into a five-frame static sequence, which is compressed into two latent frames by the temporal VAE, and use the first decoded frame as the final output.
As shown in Figure~\ref{fig:sequence_ablation_img}, this simple sequence construction recovers finer structural details than single-frame input.
We adopt five-frame replication for image training.

\vspace{-2mm}
\section{Conclusion}
\label{sec:conclusion}
\vspace{-1mm}

We presented \textbf{VitronEdit}, a unified framework for image, video, and 3D scene editing, together with geometry-aware post-training for improving structural consistency.
Our experiments reveal complementary cross-modal transfer: image and video training primarily contributes editing semantics to 3D, while limited 3D supervision mainly strengthens multi-view consistency.
We further find that restricting geometric supervision to 3D yields the best overall performance, with modest gains also transferring to image and video editing through the shared backbone.
VitronEdit thus retains competitive editing capabilities across modalities while achieving its clearest advantage in multi-view 3D consistency.

{
    \small
    \bibliographystyle{plainnat}
    \bibliography{main}
}

\clearpage

\appendix

\section{Implementation Details}
\label{app:training}
Table~\ref{tab:training_config} summarizes the training schedule
and hyperparameters.
Both training stages use 360p resolution and run on 8 NVIDIA H200 GPUs.
The LoRA rank and alpha are set to 32.
During Stage 1, we update only the DiT LoRA and modality tokens added to the MLLM vocabulary.
We reproduce the geometry extractor using the open-source code of VIST3A~\citep{go2026textto3d}, with the HunyuanVideo-1.5 VAE~\citep{wu2025hunyuanvideo}.
During stage 2, the DiT LoRA is updated, while the geometry extractor and modality tokens remain frozen.
The non-editing DL3DV sequences in Table~\ref{tab:training_data} are used primarily during geometry-aware post-training, with the same sequence serving as both source and target, to provide geometric supervision under large viewpoint changes.

\paragraph{Sequence construction.}
For the constructed 72-frame multiview editing pairs, we select 26 views in Stage 1 and 17 views in Stage 2, using identical frame indices for the source and target sequences.
We retain the first selected view once and repeat each subsequent view four times, yielding $1+4(26-1)=101$ pixel frames in Stage 1 and $1+4(17-1)=65$ pixel frames in Stage 2.
The temporal VAE compresses these sequences into 26 and 17 latent frames, respectively.
For images, Phase 1 uses a single frame; five-frame static replication is introduced in Phase 2 and retained in Phase 3 and Stage 2.

\begin{table}[!htbp]
\centering
\caption{
Training schedule and hyperparameters.
I, V, and 3D denote image, video, and 3D multi-view editing,
respectively. A dash indicates an inactive modality.
}
\label{tab:training_config}
\fontsize{9}{10}\selectfont
\setlength{\tabcolsep}{4mm}
\renewcommand{\arraystretch}{1.1}
\begin{tabular}{@{}lcccc@{}}
\toprule
\multirow[c]{2}{*}{\textbf{Setting}}
& \multicolumn{3}{c}{\textbf{Stage 1: progressive training}}
& \textbf{Stage 2} \\
\cmidrule(lr){2-4} \cmidrule(l){5-5}
& \textbf{Ph. 1 (I)}
& \textbf{Ph. 2 (I+V)}
& \textbf{Ph. 3 (I+V+3D)}
& \textbf{Geometry-aware} \\
\midrule
Training steps
& 10K & 7K & 3K & 500 \\
\midrule
Learning rate (LoRA)
& $5\times10^{-5}$
& $5\times10^{-5}$
& $2\times10^{-5}$
& $1\times10^{-5}$ \\
Learning rate (modality tokens)
& $1\times10^{-5}$
& $1\times10^{-5}$
& $5\times10^{-6}$
& Frozen \\
\midrule
Effective batch size (image)
& 200 & 480 & 480 & 128 \\
Effective batch size (video)
& / & 120 & 120 & 32 \\
Effective batch size (3D)
& / & / & 120 & 32 \\
\midrule
Sequence length (image)
& 1 & 5 & 5 & 5 \\
Sequence length (video)
& / & 65 & 65 & 65 \\
Sequence length (3D)
& / & / & 101 & 65 \\
\bottomrule
\end{tabular}
\end{table}

\begin{table}[!htbp]
\centering
\caption{Training data used for image, video, and 3D multi-view editing.}
\label{tab:training_data}
\fontsize{9}{9}\selectfont
\setlength{\tabcolsep}{6.5mm}
\renewcommand{\arraystretch}{1.1}
\begin{tabular}{@{}lllc@{}}
\toprule
\textbf{Modality} &
\textbf{Data Source} &
\textbf{Task Type} &
\textbf{Samples} \\
\midrule

\multirow[c]{3}{*}{Image}
& AnyEdit~\citep{yu2025anyedit}
& Instruct-based Editing
& \multirow[c]{3}{*}{$\sim$2M} \\

& NHR-Edit~\citep{kuprashevich2026nohumansrequired}
& Instruct-based Editing
& \\

& AnyInsertion~\citep{song2026insertanything}
& Reference-based Editing
& \\

\midrule

\multirow[c]{2}{*}{Video}
& OpenVE~\citep{he2026openve}
& Instruct-based Editing
& \multirow[c]{2}{*}{$\sim$500K} \\

& Kiwi-Edit~\citep{lin2026kiwiedit}
& Reference-based Editing
& \\

\midrule

\multirow[c]{3}{*}{3D}
& Self-built scene editing dataset
& Instruct-based Editing
& $\sim$8K \\

& Nano3D~\citep{ye2026nano3d}
& Instruct-based Editing
& $\sim$20K \\

& DL3DV~\citep{ling2024dl3dv}
& Non-editing
& $\sim$2K \\

\bottomrule
\end{tabular}
\end{table}

\section{Geometry Loss Details}
\label{app:geo_loss}

\paragraph{Online generation for geometric supervision.}
Geometry-aware post-training includes online inference with the editing model to obtain the predicted edited latent sequence, denoted by $\hat z_{\mathrm{edit}}$.
During online generation, we use 40--50 sampling steps and retain gradients through two randomly selected steps.
The frozen latent-native geometry extractor processes this generated latent sequence and the corresponding target latent sequence to predict their depth maps and camera poses.
The geometric losses below compare these two outputs; the target branch is evaluated without gradients.

\subsection{Depth Loss}
\label{app:depth_loss}

\paragraph{Normalization.}
Predicted and target geometries are reconstructed independently and may
differ in their overall scale. We normalize each reconstruction by its
own mean depth:
\begin{equation}
d^b =
\frac{1}{NHW}
\sum_{i=0}^{N-1}
\sum_{u=1}^{H}
\sum_{v=1}^{W}
D_i^b(u,v),
\qquad
b \in \{\mathrm{pred}, \mathrm{tgt}\},
\end{equation}
where $N$ is the number of supervised frames and $H,W$ are the depth-map dimensions. Sharing this factor across the sequence removes
a common multiplicative scale while preserving relative depth scales between frames. 
Independent per-frame normalization would instead remove these inter-frame scale differences. 
We do not subtract a depth offset, since additive shifts change depth ratios, and centering would remove sensitivity to these changes.

We retain gradients through the predicted mean depth, so differentiation accounts for both the depth values and the normalization factor.
Consequently, the gradient has zero component along the overall scale direction, consistent with the invariance of normalized depth to a common multiplicative rescaling.

\paragraph{Supervision setup.}
The predicted and target branches use matching supervision positions. For a 65-frame video sequence, the 17 supervised positions correspond to pixel-frame indices $0,4,\ldots,64$. For a multiview sequence, the 65-frame replicated input represents 17 selected views, and geometric supervision compares the corresponding 17 positions in the predicted and target branches. 
For image editing, only the first latent frame is passed to the geometry extractor, and normalization uses the resulting depth map. 
The target branch uses the latent given by the VAE posterior mode to avoid randomness from posterior sampling and is evaluated without gradients.

\subsection{Camera Loss}
\label{app:camera_loss}

Camera supervision requires a common coordinate convention and comparable reconstruction scales. 
We first align the coordinate systems, then compare rotations and depth-normalized translations. 
Translation supervision combines local displacement and global position constraints.

\paragraph{Coordinate alignment.}
Independent reconstructions may use different world coordinate systems.
Direct pose comparison would therefore include coordinate-frame differences in the trajectory error. We express each trajectory in its own first-camera coordinate system:
\begin{equation}
\bar{T}_i^{\mathrm{pred}}
=
\operatorname{sg}(T_0^{\mathrm{pred}})^{-1}
T_i^{\mathrm{pred}},
\qquad
\bar{T}_i^{\mathrm{tgt}}
=
(T_0^{\mathrm{tgt}})^{-1}
T_i^{\mathrm{tgt}},
\end{equation}
where $T_i$ is a camera-to-world pose and $\operatorname{sg}$ denotes stop-gradient. 
The reference transform is treated as fixed during backpropagation.

We check the extractor outputs to determine whether this alignment is already enforced. 
The median angular deviation of the target first-camera rotation from the identity is $0.73^\circ$ for 3D and $0.23^\circ$ for video, indicating that explicit alignment is needed to enforce this convention. 
We also test invariance by left-multiplying all poses of one trajectory by the same rigid transformation $Q$, while keeping the other trajectory unchanged. 
The aligned camera loss remains unchanged, consistent with
\begin{equation}
(QT_0)^{-1}(QT_i) = T_0^{-1}T_i.
\end{equation}

\FloatBarrier

\paragraph{Rotation distance.}
After alignment, we compare rotations using the scaled chordal distance.
For a relative rotation angle $\theta \in [0,\pi]$,
\begin{equation}
\frac{
\|\bar{R}_i^{\mathrm{pred}}-\bar{R}_i^{\mathrm{tgt}}\|_F
}{\sqrt{2}}
=
2\sin(\theta/2).
\end{equation}
This distance increases monotonically with $\theta$, approximates the angular error at small angles, and can be computed directly from rotation matrices without evaluating an inverse cosine. 
The aligned first-camera rotation is the identity, so the rotation loss is evaluated only on the remaining frames.

\paragraph{Translation scale normalization.}
Camera translations must also be normalized to account for differences in reconstruction scale. 
Using the mean inter-frame displacement as the denominator introduces an inverse dependence on camera-motion magnitude into the normalization derivative. 
This can amplify gradients when camera motion is small.

We instead use mean depth, which shares the reconstruction scale of camera translation without using motion magnitude as the denominator:
\begin{equation}
p_i^{\mathrm{pred}}
=
\frac{\bar{t}_i^{\mathrm{pred}}}
{\operatorname{sg}(d^{\mathrm{pred}})},
\qquad
p_i^{\mathrm{tgt}}
=
\frac{\bar{t}_i^{\mathrm{tgt}}}
{d^{\mathrm{tgt}}}.
\end{equation}
This removes the common reconstruction scale and avoids the explicit inverse-motion factor. 
The depth denominator is treated as a fixed scale reference during backpropagation, so translation supervision does not propagate through the depth scale.

\paragraph{Local displacement and global position.}
Local displacement errors can either accumulate or cancel along a trajectory. 
Their magnitudes alone therefore do not determine the global position error. We complement adjacent-frame displacement supervision with a position anchor relative to the first frame.
Writing the normalized position error as
\begin{equation}
e_i = p_i^{\mathrm{pred}} - p_i^{\mathrm{tgt}},
\end{equation}
the local residual is $e_i-e_{i-1}$, whereas the anchor residual is
$e_i$ after alignment.

An analytical comparison illustrates the distinction. 
We fix the target trajectory and construct two predicted error patterns. 
One adds the same error vector $\delta$ at every step. 
The other alternates the error increment between $\delta$ and $-\delta$. Let $M=N-1$ be even and set $e_0=0$. 
As shown in Table~\ref{tab:translation_error_patterns}, both patterns have the same local loss because SmoothL1 is symmetric with respect to the residual sign. 
Their anchor losses differ because only the first pattern accumulates position error.

\begin{table}[!htbp]
\centering
\caption{
Analytical comparison of translation error patterns. Here, $\phi$ denotes SmoothL1 applied to each coordinate and averaged over the three coordinates. Equal local losses can correspond to different accumulated position errors.
}
\label{tab:translation_error_patterns}
\fontsize{9}{10}\selectfont
\setlength{\tabcolsep}{4mm}
\renewcommand{\arraystretch}{1.1}
\begin{tabular}{@{}lcc@{}}
\toprule
\textbf{Quantity} & \textbf{Same-direction accumulation} & \textbf{Alternating errors} \\
\midrule
Position error $e_i$
& $i\delta$
& $\frac{1-(-1)^i}{2}\delta$ \\[3pt]
Displacement error
& $\delta$
& $(-1)^{i+1}\delta$ \\[3pt]
Mean local loss
& $\phi(\delta)$
& $\phi(\delta)$ \\[3pt]
Mean anchor loss
& $\frac{1}{M}\sum_{i=1}^{M}\phi(i\delta)$
& $\frac{1}{2}\phi(\delta)$ \\[3pt]
Final position error
& $M\delta$
& $0$ \\
\bottomrule
\end{tabular}
\end{table}

We further examine whether model predictions exhibit larger position errors later in the sequence. 
We select 10 source samples from each of the 3D editing, OpenVE, and Kiwi-Edit datasets and generate predictions from three checkpoints obtained during Stage 1 training, using 50 sampling steps and three random seeds per checkpoint. 
This yields 90 prediction-target pairs for 3D and 180 for video across checkpoints and seeds. 
For each trajectory, we divide the non-reference frames into two halves and compute the mean position error $\|e_i\|_2$ in each half. 
The median ratio of second-half to first-half error is $2.23$ for 3D and $1.43$ for video. 
These larger errors in the second half further motivate explicit position supervision alongside the local displacement loss.

\subsection{Hyperparameters}
\label{app:geo_hyperparameters}

We calibrate the geometry weights in three steps.
First, using synthetic trajectories, we balance the local
displacement and global position terms based on their gradient
norms with respect to predicted translations.
Second, we use offline gradients with respect to geometry
outputs and online gradients on trainable DiT LoRA parameters
to balance the depth, translation, and rotation components.
Finally, we calibrate the overall geometry strength separately
for each modality, targeting a median ratio of approximately
$5\%$--$10\%$ between the weighted geometry and flow-matching
gradient norms on trainable DiT LoRA parameters.
Weights are adjusted between training runs and remain fixed
within each run.

Table~\ref{tab:geo_hyperparameters} summarizes the SmoothL1
thresholds and final effective weights, including all component
and modality multipliers.
The flow-matching weight is 1.
Our final model adopts Geo-3D, enabling geometric supervision
only for 3D samples; image and video geometry weights are
reported for the corresponding ablation settings.
In the ablations, we disable the specified geometry components
or modality-specific supervision while keeping all remaining
weights unchanged, without renormalization.

\begin{table}[!htbp]
\centering
\caption{
SmoothL1 thresholds and effective geometry loss weights.
Image and video weights apply only to ablation settings.
A dash indicates an inapplicable setting.
}
\label{tab:geo_hyperparameters}
\fontsize{9}{10}\selectfont
\setlength{\tabcolsep}{6mm}
\renewcommand{\arraystretch}{1.1}
\begin{tabular}{@{}llccc@{}}
\toprule
\textbf{Category} & \textbf{Parameter}
& \textbf{Image} & \textbf{Video} & \textbf{3D} \\
\midrule
\multirow{2}{*}{SmoothL1 thresholds}
    & Depth $\beta_d$         & 0.02 & 0.02  & 0.02  \\
    & Translation $\beta_t$   & --   & 0.001 & 0.001 \\
\midrule
\multirow{2}{*}{Translation weights}
    & Local $\lambda_{\mathrm{rel}}$  & -- & 1.00 & 1.00 \\
    & Anchor $\lambda_{\mathrm{abs}}$ & -- & 1.30 & 1.30 \\
\midrule
\multirow{3}{*}{Effective loss weights}
    & Depth $\lambda_D$       & 0.47 & 0.13 & 0.45 \\
    & Translation $\lambda_T$ & --   & 0.73 & 2.62 \\
    & Rotation $\lambda_R$    & --   & 0.52 & 1.85 \\
\bottomrule
\end{tabular}
\end{table}

\section{Evaluation Protocols}
\label{app:eval-protocols}

\subsection{Image Editing}

\subsubsection{Benchmark}

We evaluate image editing on the PIE-Bench~\citep{ju2024pnp} and the reference-based editing subset of ICE-Bench~\citep{pan2025icebench}.
For evaluation, the edited images produced by the baseline methods are resized to match the dimensions of their corresponding input images.

The PIE-Bench dataset contains 700 samples across ten editing categories: random editing (140); object replacement, addition, and removal (80 each); content, pose, color, and material changes (40 each); and background and style changes (80 each).
Each sample provides a source image, an editing instruction, source and target descriptions, and an editing-region mask.
The reference-based editing subset of ICE-Bench contains 518 samples: face swapping (95), reference-guided style editing (186), subject-guided inpainting (140), and virtual try-on (97).
Each sample provides a source image, an editing instruction, a target description, and a reference image.

\subsubsection{Metrics}

We evaluate image editing along two dimensions: \emph{Consistency} and \emph{CLIP Semantics}.
Consistency measures how well the edited image preserves the overall structure of the source image and its unedited regions.
It includes Structure Distance and PSNR, MSE, LPIPS, and SSIM computed on the unedited regions.
CLIP Semantics measures the alignment of the edited image with the target description and the reference image.
It includes the similarity between the whole edited image and the target description (Whole), between the edited region and the target description (Edited), and between the edited image and the reference image (Reference).
The Reference metric is evaluated only on the ICE-Bench reference-editing samples.

Let $I_s$ and $I_e$ denote the source and edited images, respectively.
Let $M$ be the binary mask of the edited region and $U=1-M$ the mask of the unedited region.
For the unedited-region metrics, we set the edited regions in both images to zero, yielding $U\odot I_s$ and $U\odot I_e$.
With image intensities normalized to $[0,1]$, MSE is computed over these masked images, and PSNR is derived from it:
\begin{equation}
\text{MSE}
=\frac{1}{3HW}\sum_{p=1}^{HW}\sum_{c=1}^{3}
\bigl[U_p(I_{s,p,c}-I_{e,p,c})\bigr]^2,
\qquad
\text{PSNR}=10\log_{10}\frac{1}{\text{MSE}}.
\end{equation}
SSIM and LPIPS are likewise computed on the two masked images; LPIPS uses a SqueezeNet feature network.
All four metrics are computed over the full image dimensions, without normalizing by the area of the unedited region.

Structure Distance measures the overall structural difference between the source and edited images.
We extract key features from the final layer of DINO ViT-B/8, form a pairwise cosine-similarity matrix over the tokens of each image, and compute the mean squared difference between the two matrices:
\begin{equation}
S(I)_{ab}
=\frac{\langle k_a(I),k_b(I)\rangle}
{\|k_a(I)\|\,\|k_b(I)\|},
\qquad
\text{Structure Distance}(I_s,I_e)
=\frac{1}{T^2}\sum_{a,b=1}^{T}
\bigl[S(I_s)_{ab}-S(I_e)_{ab}\bigr]^2,
\end{equation}
where $T$ is the number of tokens and $k_a(I)$ is the key feature of token $a$ in image $I$.

For CLIP Semantics, we use CLIP ViT-L/14 and compute the cosine similarity between features, clipped below at zero.
Let $f_I$ and $f_T$ denote the image and text encoders, $t$ the target description, and $I_r$ the reference image.
The three scores are
\begin{equation}
\begin{aligned}
\text{Whole}&=\max\!\left\{0,\cos\bigl(f_I(I_e),f_T(t)\bigr)\right\},\\
\text{Edited}&=\max\!\left\{0,\cos\bigl(f_I(M\odot I_e),f_T(t)\bigr)\right\},\\
\text{Reference}&=\max\!\left\{0,\cos\bigl(f_I(I_e),f_I(I_r)\bigr)\right\}.
\end{aligned}
\end{equation}
For Edited, pixels outside the edited region are set to zero before encoding the full image.
Reference compares the full edited image with the reference image.
Each metric is computed per sample and then averaged over the samples for which it is available.
For the combined instruction- and reference-based evaluation, each shared metric is aggregated by weighting the two subset means by their respective numbers of valid samples for that metric. Reference is computed only on the reference-based subset.

\subsection{Video Editing}

\subsubsection{Benchmark}

We evaluate video editing on FiVE-Bench~\citep{li2025fivebench} and the reference-based editing subset of VIE-Bench~\citep{mou2025instructx}.
Our FiVE-Bench evaluation set comprises 419 editing instances from 100 distinct source videos: rigid object replacement (100), non-rigid object replacement (100), color change (100), material change (100), object addition (9), and object removal (10).
The VIE-Bench reference-based editing subset contains 20 instances: reference-guided object addition and object replacement (10 each).
Each instance has a source video and a reference image.

To accommodate the input frame requirements of the baseline methods, we uniformly sample 65 frames from each source video and set the frame rate to 24 fps.
Source frames and masks are aligned by temporal index, and all methods are evaluated on output frames.
Before computing frame-level metrics, we resize the source frames, edited frames, and masks to a common resolution according to the source video's aspect ratio: $624\times352$ (landscape), $480\times480$ (square), or $352\times624$ (portrait).

\subsubsection{Metrics}

We evaluate video editing along three dimensions: \emph{Consistency}, \emph{CLIP Semantics}, and \emph{Temporal Quality}.
The evaluation uses both frame-level image metrics and video-level metrics: PSNR, LPIPS, SSIM, and CLIP similarities are computed for each frame and averaged over the valid frames of each video, whereas MFS, MFS-Edit, and DOVER~\citep{wu2023dover} are computed on the complete video.
Consistency measures the preservation of unedited content using PSNR, LPIPS, and SSIM.
CLIP Semantics measures alignment with the target description and reference image using Whole, Edited, and Reference.
Temporal Quality assesses motion fidelity and video quality using MFS, MFS-Edit, and DOVER.
Reference is evaluated only on the VIE-Bench reference-based editing instances.

The frame-level PSNR, LPIPS, SSIM, and CLIP metrics follow the image-editing definitions above.
FiVE-Bench provides frame-wise editing masks.
For VIE-Bench object replacement, we use manually verified seed boxes to prompt SAM~2~\citep{ravi2024sam2} to segment and propagate the source object through the video.
For object addition, we use full-frame masks; these instances have no unedited region and are excluded from PSNR, LPIPS, and SSIM.
For the CLIP metrics, the target description is taken from \texttt{target\_prompt} when available and otherwise from the editing instruction.
Whole compares the full edited frame with this description, whereas Edited first zeros pixels outside the editing mask.
Reference compares the full edited frame with the full reference image.

For Temporal Quality, MFS uses the FiVE-Bench implementation of CoTracker~\citep{karaev2024cotracker} to track points in the source and edited videos and compare the directions of their displacements between adjacent frames.
MFS is computed over the full frame.
MFS-Edit restricts tracking to the bounding box enclosing the union of all nonempty frame-wise editing masks, yielding a fixed region across frames.
DOVER evaluates each edited video and outputs a video quality score.
For the combined instruction- and reference-based evaluation, each shared metric is aggregated by weighting the two subset means by their respective numbers of valid editing instances for that metric. Each editing instance contributes one score after any frame-level averaging; Reference is computed only on the reference-based subset. Instances excluded from a metric, such as reference-based object addition for PSNR, LPIPS, and SSIM, are excluded from its aggregation weights.

\subsection{3D Editing}

\subsubsection{Benchmark}

Our 3D editing evaluation follows the scene and editing-instruction setup of EditSplat~\citep{lee2025editsplat}.
The benchmark comprises eight scenes, each with two editing instructions, for a total of 16 editing tasks.
The scenes are drawn from Instruct-NeRF2NeRF~\citep{haque2023instructnerf2nerf}, Mip-NeRF 360~\citep{barron2022mipnerf360}, and BlendedMVS~\citep{yao2020blendedmvs}.
We use the corresponding benchmark annotations for the editing instructions and the source and target text descriptions.

To standardize the inputs and rendering conditions, we first reconstruct an unedited 3D Gaussian Splatting (3DGS) representation~\citep{kerbl2023gaussians} for each source scene from the same images and camera data, training it for 30,000 iterations.

\paragraph{Camera trajectory and source views.}
For each scene, we interpolate 26 camera poses along a continuous segment of the dataset's calibrated training-camera trajectory.
Only the two endpoint cameras coincide with the original dataset cameras; the held-out test cameras are not used.
We render the unedited 3DGS at these poses to form the source sequence.
The angular spans are $119^\circ$ for bear and bicycle, $70^\circ$ for bonsai, $71^\circ$ for face and person, $53^\circ$ for fangzhou, $119^\circ$ for garden, and $122^\circ$ for stone\_horse.

\paragraph{Joint editing and scene reconstruction.}
VitronEdit uses the image-based reconstruction route for all reported 3D evaluations, with direct latent-to-3DGS decoding disabled.
The 26 source views are jointly edited using 50 denoising steps.
Keeping the first view once and repeating each subsequent view four times produces 101 pixel frames, corresponding to 26 latent frames.
After VAE decoding, we extract the 26 corresponding edited images.
The generated images have a long-side resolution of 448 pixels and are resized with Lanczos interpolation to the calibrated image resolution ($480\times480$ or $624\times352$) before reconstruction.

We initialize a new vanilla 3DGS from the original scene's SfM sparse points, without reusing the unedited scene's Gaussian parameters.
We train it on the 26 edited images and their associated calibrated cameras for 30,000 iterations using $0.8L_1+0.2L_{\mathrm{D\text{-}SSIM}}$.
Densification runs from iteration 500 to 15,000 at intervals of 100 iterations, with a gradient threshold of $2\times10^{-4}$.
Opacity is reset every 3,000 iterations, and the spherical-harmonic degree is 3.
The reconstruction uses the same dataset-derived camera poses as source rendering, rather than camera predictions from the geometry extractor.

\paragraph{Evaluation views.}
We select eight views with zero-based indices $\{0,4,7,11,14,18,21,25\}$ from the 26-view trajectory.
All methods render their edited scenes from these same cameras, yielding 16 tasks and 128 rendered images per method.
These eight views are a subset of the views used to reconstruct the edited 3DGS; evaluation therefore measures performance at reconstruction training views, not held-out novel views.
All reported 3D metrics are computed on the reconstructed scene renderings and evaluate the complete editing-and-reconstruction pipeline.
When computing the editing direction, we use the rendering of the unedited 3DGS from the same camera as the source image.
We follow the EditSplat task set and recompute the scores of all methods under this unified view setting.

\subsubsection{Metrics}

We evaluate 3D editing in terms of semantic accuracy and multi-view consistency using four metrics.
CLIPDir and CLIPSim follow EditSplat~\citep{lee2025editsplat}.
We additionally use 3D Consistency, adapted from the editing-consistency metric of Group Editing~\citep{ma2026groupediting}, and Ph-Loss, the photometric reprojection loss used by RL3DEdit~\citep{wang2026rl3dedit}.

Let $I_s^v$ and $I_e^v$ denote the unedited and edited 3DGS renderings at view $v$, respectively, and let $t_s$ and $t_e$ be the source and target text descriptions.
With CLIP image and text encoders $f_I$ and $f_T$, respectively, CLIPDir measures the alignment between image and text change directions, whereas CLIPSim measures the similarity between the edited rendering and target description:
\begin{equation}
\begin{aligned}
\text{CLIPDir}^{v}
&=\cos\bigl(f_I(I_e^v)-f_I(I_s^v),\,
             f_T(t_e)-f_T(t_s)\bigr),\\
\text{CLIPSim}^{v}
&=\cos\bigl(f_I(I_e^v),f_T(t_e)\bigr).
\end{aligned}
\end{equation}
Here $I_s^v$ is rendered from the unedited 3DGS at the same camera view.
Both metrics are computed at each of the eight evaluation views and averaged within each editing task; higher scores are better.

For 3D Consistency, we encode the source and edited renderings at each view using CLIP ViT-B/32 with standard preprocessing: resizing the short side to 224 pixels and center-cropping to $224\times224$.
Let $g(I)$ denote the L2-normalized image embedding. The edit direction at view $v$ is
\begin{equation}
d^v = \frac{g(I_e^v)-g(I_s^v)}{\|g(I_e^v)-g(I_s^v)\|_2}.
\end{equation}
We compute the mean cosine similarity of these edit directions over all 28 unordered pairs of the eight views, then average the task-level scores over the 16 editing tasks.
Higher scores indicate more consistent edit directions across views.

For Ph-Loss, we use VGGT-1B~\citep{wang2025vggt} to jointly predict depth and camera poses from the eight edited renderings, using a nominal input long side of 518 pixels with dimensions adjusted to multiples of 14.
We align the predicted depth to the dataset camera scale using a single scale factor per editing task.
Specifically, for each of the 28 view pairs, we divide the distance between the dataset camera centers by the distance between the corresponding VGGT-predicted camera centers, take the median of these ratios, and multiply all eight predicted depth maps by this factor.
The predicted camera poses are used only for scale estimation; reprojection uses the dataset camera poses and intrinsics.

We evaluate the seven adjacent-view pairs using forward splatting from each view to the next.
At pixels whose reprojected depth differs from the target-view depth by less than $5\%$ of the median valid target-view depth, we compute $0.15L_1+0.85(1-\mathrm{SSIM})/2$, with a $3\times3$ SSIM window.
Pairs with no valid pixels are excluded.
We average the valid pairwise errors within each editing task, then average the valid task-level scores over the 16-task benchmark. Lower Ph-Loss indicates lower photometric reprojection error in the depth-consistent regions.

\section{Construction of Multi-view Editing Data}
\label{app:multiview_data}

We construct multi-view editing data from videos of static indoor scenes recorded by a moving camera.
Frames from each video show the same scene from different viewpoints.
Each sample contains a source sequence, an editing instruction, and an edited sequence.
The pipeline consists of source sequence preparation, instruction construction, and paired edit generation.

\subsection{Source Sequence Preparation}
\label{app:multiview_source}

We use the RealEstate10K~\citep{zhou2018stereo} subset of RealCam-Vid~\citep{zheng2025realcam} and retain clips at a resolution of $1280 \times 720$.
For each clip with at least 72 frames, we uniformly sample 72 frames over its full duration.
This preserves coverage of the original camera trajectory.
We encode the sampled frames as a 3-second video at 24 fps.

We filter the sampled sequences using two measures, computed on grayscale frames $I_t$ resized to a width of 640 pixels.
We compute the Laplacian variance for each frame,
\begin{equation}
v_t = \operatorname{Var}[\nabla^2 I_t],
\end{equation}
where $\nabla^2$ is the discrete Laplacian.
We require a median of at least 90 and a 10th percentile of at least 40.
No more than 35\% of frames may have a variance below 45.

We also compute dense optical flow between consecutive frames.
Let $m_t$ denote the mean flow magnitude for each frame pair.
We discard sequences whose coefficient of variation $J = \sigma(m)/\mu(m)$ exceeds 0.85.
This measure captures variation in inter-frame flow magnitude.

\subsection{Target Selection and Instruction Construction}
\label{app:multiview_instructions}

\paragraph{Target selection.}
We use Qwen3-VL-30B-A3B~\citep{bai2025qwen3vl} to describe the objects in each sequence and estimate how reliably they can be identified across views.
We select one target object for local editing.
Local operations can modify, move, replace, or remove this object.
For object addition, it serves as a reference for placing the new object.

Candidate objects must satisfy the following conditions, based on the model's predictions:
\begin{itemize}
    \item Recognition confidence of at least 0.45.
    \item Normalized temporal span of at least 0.55.
    \item Tracking stability of at least 0.45.
    \item Center proximity of at least 0.30.
    \item At least two distinguishing visual cues.
    \item Identification as a single instance.
    \item Instance-count confidence of at least 0.60.
\end{itemize}

The normalized temporal span measures the interval between the object's first and last appearance relative to the sequence duration.
Tracking stability indicates how reliably the model identifies the same object across frames.
Center proximity indicates how close the object is to the image center.

We rank eligible objects using a weighted score.
The score favors salient objects that are identified with high confidence.
It also favors objects that remain visible over time, are tracked consistently, and lie near the image center.
We penalize objects that may be confused with others and select the highest-ranked candidate.
Sequences without an eligible target are discarded.

\paragraph{Editing operations.}
We consider ten editing operations:
\begin{itemize}
    \item \textbf{Local edits.} Object addition, removal, replacement, and movement.
    Local edits also include changes to the target's color, material, or texture.

    \item \textbf{Global edits.} Background change, tone transfer, and style change.
\end{itemize}

We use Qwen3.5-27B~\citep{qwen2026qwen35} to generate four instruction groups for each scene.
Each group describes one editing intent using three English paraphrases.
The target and operation remain the same within a group.
For global edits, the prompt provides six candidates from the corresponding background, tone, or style pool.

\paragraph{Instruction validation and selection.}
We check whether each instruction describes a specific, observable edit and matches its assigned operation.
The checks reject:
\begin{itemize}
    \item Ambiguous references or references to multiple target objects.
    \item Internal object identifiers.
    \item Viewpoint-dependent directional terms.
    \item Requests that conflict with the assigned operation.
    For example, a removal instruction must not request a replacement.
    A replacement instruction must name both the original object and its replacement.
\end{itemize}

Each instruction is limited to 30 English words and 180 characters during post-processing.
A group is discarded if it contains an invalid instruction or an instruction that duplicates another in the same scene.
Each scene retains up to three valid groups, with one paraphrase selected from each group.
We then choose one instruction per scene to bring the overall operation distribution closer to the desired proportions.

\subsection{Paired Edit Generation}
\label{app:multiview_generation}

We provide Kling-O1~\citep{team2025kling} with the source sequence and selected instruction to generate the edited sequence.
The prompt asks the model to follow the original camera trajectory and perspective, while preserving scene layout, depth relationships, and content except where changes are required by the editing instruction.

The final dataset contains 8,075 editing triplets.
Table~\ref{tab:multiview_edit_distribution} reports the operation distribution.
For reporting, material and texture changes are grouped, as are global style and tone changes.

\begin{table}[!htbp]
\centering
\caption{
Distribution of editing operations in the final dataset.
Percentages may not sum to 100 due to rounding.
}
\label{tab:multiview_edit_distribution}
\fontsize{9}{10}\selectfont
\setlength{\tabcolsep}{8mm}
\renewcommand{\arraystretch}{1.1}
\begin{tabular}{@{}lcc@{}}
\toprule
\textbf{Editing operation} & \textbf{Count} & \textbf{Share} \\
\midrule
Material or texture change & 2,368 & 29.3\% \\
Color change & 1,433 & 17.7\% \\
Global style or tone change & 1,005 & 12.4\% \\
Object replacement & 829 & 10.3\% \\
Object removal & 809 & 10.0\% \\
Object addition & 745 & 9.2\% \\
Background replacement & 711 & 8.8\% \\
Object movement & 175 & 2.2\% \\
\midrule
Total & 8,075 & 100.0\% \\
\bottomrule
\end{tabular}
\end{table}

\subsection{Prompts}
\label{app:multiview_prompts}

We give the two vision-language-model prompts used above.


\begin{promptbox}{Scene Description and Target Extraction Prompt}
You are a video-understanding model.
Given multiple frames, output strict JSON; no Markdown, no extra explanation.
Goal: extract structured features that support precise single-target editing while avoiding ambiguous references.

\textbf{Key principles:}
\begin{itemize}[leftmargin=1.2em,itemsep=1pt,topsep=2pt]
\item Object ids must be stable and unique (e.g., obj\_001, obj\_002).
\item Every editable foreground object must be given 2-4 non-positional disambiguating cues (clothing, accessories, texture, pose, etc.).
\item Do not use left/right/front/back/near/far as the primary basis for disambiguation.
\item Record each object's temporal persistence and centrality, for later instruction selection.
\item If a judgment cannot be made reliably, do not fabricate a value; use null or lower the confidence.
\end{itemize}

\textbf{Required output fields:}
\begin{enumerate}
\item \texttt{objects}: array of key foreground objects, each with: id, name, saliency (0-1), confidence (0-1); attributes: \{gender, age\_group, clothing, accessories, colors, material, texture, pose, action\_state\}; quantity: \{count\_estimate (>=1), count\_confidence (0-1), is\_single\_instance (true/false)\}; position: \{bbox\_norm: [x1,y1,x2,y2], center\_norm: [cx,cy], depth\_layer: foreground/mid/background\}; temporal\_presence: \{visible\_frame\_ratio (0-1), first\_seen\_norm (0-1), last\_seen\_norm (0-1), track\_stability (0-1)\}; center\_proximity (0-1); unique\_cues: [phrases distinguishing this object, at least 2, non-positional preferred].
(quantity is required; objects with count\_estimate > 1 are filtered out downstream.)
\item \texttt{scene}: scene information, including location, time, weather (if determinable).
\item \texttt{actions}: array of actions, each with subject\_id, verb, object\_id\_or\_name, confidence (0-1).
\item \texttt{global\_style}: overall visual style (e.g., realistic / cinematic / cartoon).
\item \texttt{camera}: camera motion (e.g., static/pan/tilt/zoom/handheld).
\item \texttt{ambiguity\_report}: array of groups of potentially confusable objects, each with: candidate\_ids: [...], confusion\_reason, best\_disambiguators: [non-positional preferred].
\end{enumerate}

Return only a single JSON object.
\end{promptbox}


\begin{promptbox}{Instruction Generation Prompt}
\textbf{Role.}
You are a video-editing-instruction designer.
Based on the input video's features, generate high-quality, diverse, executable editing instructions.
Generation proceeds in three steps: (1) use the preselected editing target, (2) select the editing operator, (3) write the instruction text.
The rules below are organized by these three steps.

\textbf{Output format.}
If you use extended reasoning, keep it under 300 words and do not draft the full text of each instruction there; after reasoning, you must give the complete final JSON answer.
Generate exactly 12 instructions and return strict JSON (no explanatory text).
Top-level field: \texttt{instructions}.
Each instruction has fields: \texttt{id}, \texttt{operation\_scope}, \texttt{operation}, \texttt{target}, \texttt{instruction}, \texttt{rationale}, \texttt{semantic\_group}, and \texttt{variant\_id}.
\texttt{operation\_scope} may only be \texttt{local} or \texttt{global}.

\textbf{Step 1: Use the preselected editing target}
(\texttt{local} = single-object operation / \texttt{global} = scene-level operation).
\begin{itemize}[leftmargin=1.2em,itemsep=1pt,topsep=2pt]
\item The input contains one preselected object, \texttt{objects[0]}, obtained through the upstream filtering and ranking procedure.
Do not select another object or relax the target-selection criteria.
\item Each local instruction must describe a single editing operation.
For removal, replacement, movement, and attribute changes, the preselected object is the sole editing target.
For addition, it serves as the placement reference and must remain unchanged.
\item Identify the preselected object using at least two non-positional visual cues provided in the input.
Do not invent identifying attributes.
\item Ambiguous, group, or collective references are forbidden (e.g., ``one of them,'' ``all chairs,'' or ``people'').
Refer to the preselected object as a single, identifiable instance.
\item Do not use viewpoint-dependent directions to identify the target or specify placement.
Use stable scene relations for addition and movement, such as ``on the tabletop'' or ``attached to the wall,'' and avoid camera-relative descriptions such as ``on the left side of the frame'' or ``facing the camera.''
\item Global instructions apply at the scene level and do not designate the preselected object as an individual editing target.
\end{itemize}

\textbf{Step 2: Select the editing operator.}
\begin{itemize}[leftmargin=1.2em,itemsep=1pt,topsep=2pt]
\item Local-only operators: \texttt{remove}, \texttt{replace}, \texttt{add},\texttt{move},\texttt{change\_color}, \texttt{change\_material}, \texttt{change\_texture}.
\item Global-only operators: \texttt{change\_background}, \texttt{tone\_transfer}, \texttt{style\_change}.
\end{itemize}

A local operation must be one of \texttt{remove}/\texttt{replace}/ \texttt{add}/\texttt{change\_color}/\texttt{change\_material}/\texttt{move}/ \texttt{change\_texture}; each is defined as follows:
\begin{itemize}[leftmargin=1.2em,itemsep=3pt,topsep=2pt]
\item \textbf{add:} Introduce one new object while preserving the preselected reference object.
Specify (1) the new object's concrete category (e.g., ``a small potted plant,'' not ``an object'' or ``a prop''); and (2) its placement relative to the preselected object using a stable scene relation, such as ``on the tabletop'' or ``attached to the cabinet door.''
Specify orientation only when relevant, using a fixed scene reference rather than the camera.
The placement must be physically plausible given the input scene.
\emph{Example:} ``Add a small potted plant on the \texttt{\{name\}}, leaving the table unchanged.''
\item \textbf{remove:} Delete the target object only; pure deletion, with no wording implying a replacement or new insertion (e.g., ``replace it with'' / ``instead'' / ``and add'' / ``put a new'' are all forbidden).
\emph{Example:} ``Remove the \texttt{\{name\}} from the scene while preserving the original background continuity.''
\item \textbf{move:} Reposition the target object to a different plausible location while preserving its identity, appearance, and scene consistency.
The instruction must specify the new position without changing the object itself.
\emph{Example:} ``Move the \texttt{\{name\}} closer to the wall while preserving its appearance and orientation.''
\item \textbf{replace:} Swap the target for a different object.
Before writing, first settle in your own mind exactly what is being swapped for what -- pick a replacement that is plausible for the scene and location (e.g., swapping a kitchen-island stool for a bench, or a dining chair for a sideboard -- something that fits the scene's logic, not an arbitrary mismatched object) -- then the instruction text must name both objects explicitly (``swap the specific original object for the specific new object''), neither omittable.
Vague statements that only describe a change in shape without naming the new object (e.g., ``a different-category object,'' ``an alternative design,'' ``something with a different shape'') are not acceptable, because they never say what the new object actually is.
The new object's category, shape, or silhouette must differ clearly from the original, and a pure material or color change must not be phrased with ``replace'' (use \texttt{change\_material} or \texttt{change\_color} instead).
\emph{Example:} ``Swap the \texttt{\{name\}} for a round ottoman with a distinctly different shape and structure, keeping its footprint roughly in the same spot.''
\emph{Counter-example} (not acceptable -- states a shape difference without naming the new object): ``Replace the \texttt{\{name\}} with a distinctly different shape, featuring a dark tone.''
\item \textbf{change\_color:} Change only the target object's color/hue, not its material, shape, or texture.
A specific color must be named (e.g., ``a deep walnut brown,'' ``a matte forest green''), not just ``a different color.''
\emph{Example:} ``Change the color of the \texttt{\{name\}} to a deep walnut tone with natural shading transitions.''
\item \textbf{change\_material:} Change only the target object's material, not its shape.
A specific material must be named (e.g., ``brushed metal,'' ``matte stone,'' ``polished wood''), not just ``a different material.''
\emph{Example:} ``Change the material of the \texttt{\{name\}} to brushed metal while keeping geometry unchanged.''
\item \textbf{change\_texture:} Change only the target object's surface texture/finish (smoothness, grain, pattern, etc.), not its material category, color, or shape.
A specific texture description must be given (e.g., ``a fine-grain finish,'' ``a smoother pattern'').
\emph{Example:} ``Change the surface texture of the \texttt{\{name\}} to a visibly coarse-grained finish.''
\end{itemize}

A global operation must be one of \texttt{change\_background}/\texttt{tone\_transfer}/ \texttt{style\_change}.
Background replacement may change background content while preserving foreground objects.
Tone transfer and style change must preserve scene content and spatial layout.
Each operation is defined as follows:
\begin{itemize}[leftmargin=1.2em,itemsep=3pt,topsep=2pt]
\item \textbf{change\_background:} Replace only the background and distant scenery, while preserving the identity, appearance, position, and size of foreground objects.
Name a specific background selected from the six candidates provided for the scene: \texttt{<background candidate 1>} / \dots\ / \texttt{<background candidate 6>}.
Choose the candidate that best matches the scene's content and mood; candidates outside the list are forbidden.
Do not automatically select the first candidate.
\emph{Example:} ``Replace the background with \texttt{<background candidate 1>}, keeping foreground objects unchanged.''

\item \textbf{tone\_transfer:} Change the overall color or tonal treatment (e.g., warm/cool tone, saturation, or contrast), while preserving object identities, geometry, spatial layout, and background content.
Name a specific tone selected from the six tone candidates provided for the scene, following the selection rule above.
\emph{Example:} ``Apply \texttt{<tone candidate 1>} throughout the scene while preserving its content and layout.''

\item \textbf{style\_change:} Change the overall rendering style while preserving object identities, geometry, spatial layout, and background content.
Name a specific style selected from the six style candidates provided for the scene, following the selection rule above.
\emph{Example:} ``Render the entire scene in \texttt{<style candidate 1>}, preserving its content and layout.''
\end{itemize}
The \texttt{instruction} and \texttt{target} fields of a global instruction must not contain object IDs or designate an individual foreground object as the editing target.
Object names are allowed when describing the requested background content.

\textbf{Step 3: Write the instruction text.}
\begin{itemize}[leftmargin=1.2em,itemsep=1pt,topsep=2pt]
\item For a local instruction, the \texttt{target} field must reference the preselected object's \texttt{object\_id} and include 2--4 non-positional disambiguating cues from the input.
For addition, this field identifies the placement reference object rather than the new object.
\item Target field format:
  \begin{itemize}[leftmargin=1.2em,itemsep=1pt,topsep=1pt]
  \item local: \texttt{object\_id=<id>; name=<name>; cues=<cue1|cue2|...>; temporal\_presence=<ratio>; center\_proximity=<score>}
  \item global: \texttt{scene\_region=<global>; theme=<background\_or\_tone\_or\_style>}
  \end{itemize}
\item The instruction text must not contain fragments like ``\texttt{object\_id=...}'' or ``\texttt{obj\_001}'' (object IDs may appear only in the \texttt{target} field).
\item The instruction must be a short, single sentence, no more than about 22 English words.
\item Forbidden vague wording: unverifiable phrases such as ``visually compatible alternative,'' ``suitable counterpart,'' ``context-appropriate,'' and ``subtle'' are not allowed.
\item Every instruction must describe a clear, observable outcome, such as adding, removing, replacing, or moving an object; changing its color, material, or texture; replacing the background; or applying a specific tone or style.
\item Maximize diversity across the four semantic groups through different editing operations or clearly different intended outcomes.
Within each group, use diverse wording and sentence structures while preserving the same editing intent.
The \texttt{instruction} is written in English, and \texttt{rationale} briefly explains why the edit is reasonable.
\end{itemize}

\textbf{Grouping and variant rules.}
\begin{itemize}[leftmargin=1.2em,itemsep=1pt,topsep=2pt]
\item Divide all 12 instructions into exactly 4 semantic groups (\texttt{semantic\_group} = 1, 2, 3, or 4).
\item Each group must contain exactly 3 instructions expressing different English phrasings of the same editing intent (\texttt{variant\_id} = 1, 2, or 3).
\item Within a group, \texttt{operation\_scope}, \texttt{operation}, and \texttt{target} must be identical.
The three variants must preserve all details of the intended outcome and differ only in how the instruction is phrased.
\item The 4 groups must express distinct editing intents.
Prefer different operations where appropriate; groups using the same operation must specify clearly different outcomes.
\item Every instruction must include the \texttt{semantic\_group} and \texttt{variant\_id} fields.
\end{itemize}

\textbf{Quantity requirement.}
Exactly 12 instructions must be output.

\textbf{Pre-generation self-check.}
\begin{enumerate}
\item Does every local instruction uniquely target one object?
\item Are vague and group references avoided?
\item Are viewpoint-dependent directions and camera-relative orientations avoided in target identification and placement?
\item Does every local instruction use the preselected object as the editing target, or as the placement reference for addition?
\item Does the instruction avoid vague wording and state an observable outcome?
\item Are there exactly four distinct semantic groups, each containing three paraphrases of the same editing intent?
\end{enumerate}

\textbf{Input features (JSON).} \texttt{<the stage-1 output scene-description JSON, restricted to the selected target>}
\end{promptbox}

\section{Limitation}

One limitation is that the same geometric objectives do not benefit all modalities equally, as shown by our ablation study.
Future work could explore modality-adaptive geometric supervision, particularly for videos where camera motion and object motion are entangled.
In addition, our current 3D setting focuses on multi-view editing of static scenes.
Extending the unified framework to dynamic 3D or 4D scene editing is also an interesting direction for future work.

\end{document}

%% file: math_commands.tex
\usepackage{amsmath,amsfonts,bm}

\def\eqref#1{equation~\ref{#1}}

\def\1{\bm{1}}

\DeclareMathAlphabet{\mathsfit}{\encodingdefault}{\sfdefault}{m}{sl}
\SetMathAlphabet{\mathsfit}{bold}{\encodingdefault}{\sfdefault}{bx}{n}



%% file: main.bib
@inproceedings{bai2025geovideo,
  title = {{GeoVideo: Introducing Geometric Regularization into Video Generation Model}},
  author = {Yunpeng Bai and Shaoheng Fang and Chaohui Yu and Fan Wang and Qixing Huang},
  booktitle = {Proceedings of  NeurIPS},
  year = {2025}
}

@misc{bai2025qwen25vl,
  title = {{Qwen2.5-VL Technical Report}},
  author = {Bai, Shuai and Chen, Keqin and Liu, Xuejing and Wang, Jialin and Ge, Wenbin and Song, Sibo and Dang, Kai and Wang, Peng and Wang, Shijie and Tang, Jun and Zhong, Humen and Zhu, Yuanzhi and Yang, Mingkun and Li, Zhaohai and Wan, Jianqiang and Wang, Pengfei and Ding, Wei and Fu, Zheren and Xu, Yiheng and Ye, Jiabo and Zhang, Xi and Xie, Tianbao and Cheng, Zesen and Zhang, Hang and Yang, Zhibo and Xu, Haiyang and Lin, Junyang},
  year = {2025}
}

@inproceedings{chen2024dge,
  title = {{DGE: Direct Gaussian 3D Editing by Consistent Multi-view Editing}},
  author = {Chen, Minghao and Laina, Iro and Vedaldi, Andrea},
  booktitle = {Proceedings of ECCV},
  year = {2024},
}

@inproceedings{chen2026omni3dedit,
  title = {{Omni-3DEdit: Generalized Versatile 3D Editing in One-Pass}},
  author = {Chen, Liyi and Wang, Pengfei and Zhang, Guowen and Ma, Zhiyuan and Zhang, Lei},
  booktitle = {Proceedings of CVPR},
  year = {2026}
}

@article{chen2026vino,
  title={VINO: A Unified Visual Generator with Interleaved OmniModal Context},
  author={Chen, Junyi and He, Tong and Fu, Zhoujie and Wan, Pengfei and Gai, Kun and Ye, Weicai},
  journal={arXiv preprint arXiv:2601.02358},
  year={2026}
}

@inproceedings{du2026videogpa,
  title = {{VideoGPA: Distilling Geometry Priors for 3D-Consistent Video Generation}},
  author = {Hongyang Du and Junjie Ye and Xiaoyan Cong and Runhao Li and Jingcheng Ni and Aman Agarwal and Zeqi Zhou and Zekun Li and Randall Balestriero and Yue Wang},
  booktitle = {Proceedings of the ICML},
  year = {2026}
}

@inproceedings{go2026textto3d,
  title = {{Text-to-3D by Stitching a Multi-view Reconstruction Network to a Video Generator}},
  author = {Hyojun Go and Dominik Narnhofer and Goutam Bhat and Prune Truong and Federico Tombari and Konrad Schindler},
  booktitle = {Proceedings of the ICLR},
  year = {2026}
}

@inproceedings{he2026openve,
  title = {{OpenVE-3M: A Large-Scale High-Quality Dataset for Instruction-Guided Video Editing}},
  author = {He, Haoyang and Wang, Jie and Zhang, Jiangning and Xue, Zhucun and Bu, Xingyuan and Yang, Qiangpeng and Wen, Shilei and Xie, Lei},
  booktitle = {Proceedings of the ECCV},
  year = {2026}
}

@inproceedings{haque2023instructnerf2nerf,
  title = {{Instruct-NeRF2NeRF: Editing 3D Scenes with Instructions}},
  author = {Haque, Ayaan and Tancik, Matthew and Efros, Alexei A. and Holynski, Aleksander and Kanazawa, Angjoo},
  booktitle = {Proceedings of the ICCV},
  year = {2023}
}

@inproceedings{chen2024gaussianeditor,
  title = {{GaussianEditor: Swift and Controllable 3D Editing with Gaussian Splatting}},
  author = {Chen, Yiwen and Chen, Zilong and Zhang, Chi and Wang, Feng and Yang, Xiaofeng and Wang, Yikai and Cai, Zhongang and Yang, Lei and Liu, Huaping and Lin, Guosheng},
  booktitle = {Proceedings of the CVPR},
  year = {2024}
}

@article{hartley2013rotation,
  title={Rotation averaging},
  author={Hartley, Richard and Trumpf, Jochen and Dai, Yuchao and Li, Hongdong},
  journal={International journal of computer vision},
  year={2013}
}

@inproceedings{keetha2026mapanything,
  title = {{MapAnything: Universal Feed-Forward Metric 3D Reconstruction}},
  author = {Keetha, Nikhil and M{\"u}ller, Norman and Sch{\"o}nberger, Johannes and Porzi, Lorenzo and Zhang, Yuchen and Fischer, Tobias and Knapitsch, Arno and Zauss, Duncan and Weber, Ethan and Antunes, Nelson and Luiten, Jonathon and Lopez-Antequera, Manuel and Rota Bul{\`o}, Samuel and Richardt, Christian and Ramanan, Deva and Scherer, Sebastian and Kontschieder, Peter},
  booktitle = {Proceedings of the 3DV},
  year = {2026}
}

@inproceedings{ju2026editverse,
  title = {{EditVerse: Unifying Image and Video Editing and Generation with In-Context Learning}},
  author = {Ju, Xuan and Wang, Tianyu and Zhou, Yuqian and Zhang, He and Liu, Qing and Zhao, Nanxuan and Zhang, Zhifei and Li, Yijun and Cai, Yuanhao and Liu, Shaoteng and Pakhomov, Daniil and Lin, Zhe and Kim, Soo Ye and Xu, Qiang},
  booktitle = {Proceedings of the ICLR},
  year = {2026}
}

@inproceedings{kuprashevich2026nohumansrequired,
  title = {{NoHumansRequired: Autonomous High-Quality Image Editing Triplet Mining}},
  author = {Kuprashevich, Maksim and Alekseenko, Grigorii and Tolstykh, Irina and Fedorov, Georgii and Suleimanov, Bulat and Dokholyan, Vladimir and Gordeev, Aleksandr},
  booktitle = {Proceedings of the WACV},
  year = {2026}
}

@article{kwon2024unified,
  title={Unified editing of panorama, 3d scenes, and videos through disentangled self-attention injection},
  author={Kwon, Gihyun and Park, Jangho and Ye, Jong Chul},
  journal={arXiv preprint arXiv:2405.16823},
  year={2024}
}

@inproceedings{lee2025editsplat,
  title = {{EditSplat: Multi-View Fusion and Attention-Guided Optimization for View-Consistent 3D Scene Editing with 3D Gaussian Splatting}},
  author = {Lee, Dong In and Park, Hyeongcheol and Seo, Jiyoung and Park, Eunbyung and Park, Hyunje and Baek, Ha Dam and Shin, Sangheon and Kim, Sangmin and Kim, Sangpil},
  booktitle = {Proceedings of the CVPR},
  year = {2025}
}

@inproceedings{lipman2023flow,
  title = {{Flow Matching for Generative Modeling}},
  author = {Yaron Lipman and Ricky T. Q. Chen and Heli Ben-Hamu and Maximilian Nickel and Matt Le},
  booktitle = {Proceedings of the ICLR},
  year = {2023}
}

@misc{lin2026kiwiedit,
  title = {{Kiwi-Edit: Versatile Video Editing via Instruction and Reference Guidance}},
  author = {Lin, Yiqi and Liang, Guoqiang and Zeng, Ziyun and Bai, Zechen and Chen, Yanzhe and Shou, Mike Zheng},
  year = {2026}
}

@inproceedings{ling2024dl3dv,
  title = {{DL3DV-10K: A Large-Scale Scene Dataset for Deep Learning-based 3D Vision}},
  author = {Lu Ling and Yichen Sheng and Zhi Tu and Wentian Zhao and Cheng Xin and Kun Wan and Lantao Yu and Qianyu Guo and Zixun Yu and Yawen Lu and Xuanmao Li and Xingpeng Sun and Rohan Ashok and Aniruddha Mukherjee and Hao Kang and Xiangrui Kong and Gang Hua and Tianyi Zhang and Bed{\v r}ich Bene{\v s} and Aniket Bera},
  booktitle = {Proceedings of the CVPR},
  year = {2024}
}

@article{mou2025instructx,
  title={Instructx: Towards unified visual editing with mllm guidance},
  author={Mou, Chong and Sun, Qichao and Wu, Yanze and Zhang, Pengze and Li, Xinghui and Ye, Fulong and Zhao, Songtao and He, Qian},
  journal={arXiv preprint arXiv:2510.08485},
  year={2025}
}

@article{pan2026omniweaving,
  title={OmniWeaving: Towards Unified Video Generation with Free-form Composition and Reasoning},
  author={Pan, Kaihang and Tian, Qi and Zhang, Jianwei and Kong, Weijie and Xiong, Jiangfeng and Long, Yanxin and Zhang, Shixue and Qiu, Haiyi and Wang, Tan and Lv, Zheqi and others},
  journal={arXiv preprint arXiv:2603.24458},
  year={2026}
}

@inproceedings{chen2026ViP3DE,
  title = {{Fast Multi-view Consistent 3D Editing with Video Priors}},
  author = {Chen, Liyi and Li, Ruihuang and Zhang, Guowen and Wang, Pengfei and Zhang, Lei},
  booktitle = {Proceedings of the AAAI},
  year = {2026}
}

@inproceedings{huang2025edit360,
  title = {{Edit360: 2D Image Edits to 3D Assets from Any Angle}},
  author = {Huang, Junchao and Hu, Xinting and Shi, Shaoshuai and Tian, Zhuotao and Jiang, Li},
  booktitle = {Proceedings of the ICCV},
  year = {2025}
}

@inproceedings{ma2026groupediting,
  title = {{Group Editing: Edit Multiple Images in One Go}},
  author = {Ma, Yue and Wang, Xinyu and Ma, Qianli and Wang, Qinghe and Zheng, Mingzhe and Yang, Xiangpeng and Li, Hao and Zhao, Chongbo and Ying, Jixuan and Yang, Harry and Liu, Hongyu and Chen, Qifeng},
  booktitle = {Proceedings of the CVPR},
  year = {2026}
}

@inproceedings{li2025syncnoise,
  title = {{SyncNoise: Geometrically Consistent Noise Prediction for Instruction-based 3D Editing}},
  author = {Li, Ruihuang and Chen, Liyi and Zhang, Zhengqiang and Jampani, Varun and Patel, Vishal M and Zhang, Lei},
  booktitle = {Proceedings of the AAAI},
  year = {2025}
}

@inproceedings{parihar2025depthaware,
  title = {{Zero-Shot Depth Aware Image Editing with Diffusion Models}},
  author = {Parihar, Rishubh and VS, Sachidanand and Babu, R. Venkatesh},
  booktitle = {Proceedings of the ICCV},
  year = {2025}
}

@inproceedings{brooks2023instructpix2pix,
  title = {{InstructPix2Pix: Learning To Follow Image Editing Instructions}},
  author = {Brooks, Tim and Holynski, Aleksander and Efros, Alexei A.},
  booktitle = {Proceedings of the CVPR},
  year = {2023}
}

@inproceedings{bai2026ditto,
  title = {{Scaling Instruction-Based Video Editing with a High-Quality Synthetic Dataset}},
  author = {Bai, Qingyan and Wang, Qiuyu and Ouyang, Hao and Yu, Yue and Wang, Hanlin and Wang, Wen and Cheng, Ka Leong and Ma, Shuailei and Zeng, Yanhong and Liu, Zichen and Xu, Yinghao and Shen, Yujun and Chen, Qifeng},
  booktitle = {Proceedings of the CVPR},
  year = {2026}
}

@inproceedings{wu2025insvie,
  title = {{InsViE-1M: Effective Instruction-based Video Editing with Elaborate Dataset Construction}},
  author = {Wu, Yuhui and Chen, Liyi and Li, Ruibin and Wang, Shihao and Xie, Chenxi and Zhang, Lei},
  booktitle = {Proceedings of the ICCV},
  year = {2025}
}

@inproceedings{kim2023collaborative,
  title = {{Collaborative Score Distillation for Consistent Visual Editing}},
  author = {Kim, Subin and Lee, Kyungmin and Choi, June Suk and Jeong, Jongheon and Sohn, Kihyuk and Shin, Jinwoo},
  booktitle = {Proceedings of the NeurIPS},
  year = {2023}
}

@inproceedings{yin2026vigor,
  title = {{VIGOR: VIdeo Geometry-Oriented Reward for Temporal Generative Alignment}},
  author = {Yin, Tengjiao and Shi, Jinglei and Guo, Heng and Wang, Xi},
  booktitle = {Proceedings of the ECCV},
  year = {2026}
}

@inproceedings{wang2026rl3dedit,
  title = {{Edit in 2D, Verify in 3D: Reinforcement Learning for Multi-view Consistent Scene Editing}},
  author = {Wang, Jiyuan and Lin, Chunyu and Sun, Lei and Cao, Zhi and Yin, Yuyang and Nie, Lang and Yuan, Zhenlong and Chu, Xiangxiang and Wei, Yunchao and Liao, Kang and Lin, Guosheng},
  booktitle = {Proceedings of the ECCV},
  year = {2026}
}

@inproceedings{an2026vggrpo,
  title = {{VGGRPO: Towards World-Consistent Video Generation with 4D Latent Reward}},
  author = {An, Zhaochong and Kupyn, Orest and Uscidda, Th{\'e}o and Colaco, Andrea and Ahuja, Karan and Belongie, Serge and Gonzalez-Franco, Mar and Tintore Gazulla, Marta},
  booktitle = {Proceedings of the ECCV},
  year = {2026}
}

@inproceedings{wu2024gaussctrl,
  title = {{GaussCtrl: Multi-view Consistent Text-Driven 3D Gaussian Splatting Editing}},
  author = {Wu, Jing and Bian, Jia-Wang and Li, Xinghui and Wang, Guangrun and Reid, Ian and Torr, Philip and Prisacariu, Victor Adrian},
  booktitle = {Proceedings of the ECCV},
  year = {2024}
}

@article{rao2026dense,
  title={From Dense Prediction to Visual Editing: Structured Supervision for Unified Image and Video Creation},
  author={Rao, Zhefan and Zou, Bin and Che, Haoxuan and He, Xuanhua and Choi, Chong Hou and Li, Yanheng and Liu, Rui and Chen, Qifeng},
  journal={arXiv preprint arXiv:2608.14740},
  year={2026}
}

@inproceedings{song2026insertanything,
  title = {{Insert Anything: Image Insertion via In-Context Editing in DiT}},
  author = {Song, Wensong and Jiang, Hong and Yang, Zongxin and Cheng, Zheqiao and Quan, Ruijie and Yang, Yi},
  booktitle = {Proceedings of the AAAI},
  year = {2026}
}

@inproceedings{sturm2012benchmark,
  title = {A Benchmark for the Evaluation of {RGB-D} {SLAM} Systems},
  author = {Sturm, J{\"u}rgen and Engelhard, Nikolas and Endres, Felix and Burgard, Wolfram and Cremers, Daniel},
  booktitle = {2012 IEEE/RSJ International Conference on Intelligent Robots and Systems},
  year = {2012}
}

@inproceedings{wang2024dust3r,
  title = {{DUSt3R: Geometric 3D Vision Made Easy}},
  author = {Wang, Shuzhe and Leroy, Vincent and Cabon, Yohann and Chidlovskii, Boris and Revaud, Jerome},
  booktitle = {Proceedings of the CVPR},
  year = {2024}
}

@inproceedings{wei2026univideo,
  title = {{UniVideo: Unified Understanding, Generation, and Editing for Videos}},
  author = {Cong Wei and Quande Liu and Zixuan Ye and Qiulin Wang and Xintao Wang and Pengfei Wan and Kun Gai and Wenhu Chen},
  booktitle = {Proceedings of the ICLR},
  year = {2026}
}

@article{wu2025hunyuanvideo,
  title={HunyuanVideo 1.5 Technical Report},
  author={Wu, Bing and Zou, Chang and Li, Changlin and Huang, Duojun and Yang, Fang and Tan, Hao and Peng, Jack and Wu, Jianbing and Xiong, Jiangfeng and Jiang, Jie and others},
  journal={arXiv preprint arXiv:2511.18870},
  year={2025}
}

@inproceedings{wu2026geometryforcing,
  title = {{Geometry Forcing: Marrying Video Diffusion and 3D Representation for Consistent World Modeling}},
  author = {Haoyu Wu and Diankun Wu and Tianyu He and Junliang Guo and Yang Ye and Yueqi Duan and Jiang Bian},
  booktitle = {Proceedings of the ICLR},
  year = {2026}
}

@article{yang2026context,
  title={Context unrolling in omni models},
  author={Yang, Ceyuan and Lin, Zhijie and Zhao, Yang and Xiao, Fei and He, Hao and Zhao, Qi and Deng, Chaorui and Li, Kunchang and Ding, Zihan and Guo, Yuwei and others},
  journal={arXiv preprint arXiv:2604.21921},
  year={2026}
}

@inproceedings{ye2026nano3d,
  title = {{Nano3D: A Training-Free Approach for Efficient 3D Editing Without Masks}},
  author = {Ye, Junliang and Xie, Shenghao and Zhao, Ruowen and Wang, Zhengyi and Yan, Hongyu and Zu, Wenqiang and Ma, Lei and Zhu, Jun},
  booktitle = {Proceedings of the ICLR},
  year = {2026}
}

@inproceedings{yu2025anyedit,
  title = {{AnyEdit: Mastering Unified High-Quality Image Editing for Any Idea}},
  author = {Yu, Qifan and Chow, Wei and Yue, Zhongqi and Pan, Kaihang and Wu, Yang and Wan, Xiaoyang and Li, Juncheng and Tang, Siliang and Zhang, Hanwang and Zhuang, Yueting},
  booktitle = {Proceedings of the CVPR},
  year = {2025}
}

@article{xia2025dreamve,
  title={DreamVE: Unified instruction-based image and video editing},
  author={Xia, Bin and Liu, Jiyang and Zhang, Yuechen and Peng, Bohao and Chu, Ruihang and Wang, Yitong and Wu, Xinglong and Yu, Bei and Jia, Jiaya},
  journal={arXiv preprint arXiv:2508.06080},
  year={2025}
}

@inproceedings{zhao2026tinker,
  title = {{TINKER: Diffusion's Gift to 3D--Multi-View Consistent Editing From Sparse Inputs without Per-Scene Optimization}},
  author = {Zhao, Canyu and Li, Xiaoman and Feng, Tianjian and Zhao, Zhiyue and Chen, Hao and Shen, Chunhua},
  booktitle = {Proceedings of the ICLR},
  year = {2026}
}

@misc{zhu2026jointedit3d,
  title = {{JointEdit3D: Feed-Forward 3D Scene Editing in a Unified Latent Space}},
  author = {Zhu, Xinnan and Xu, Ruijie and Ying, Jiayu and Dong, Daoguo and Xu, Jiachen and Xie, Yuan and Tan, Xin},
  year = {2026}
}

@inproceedings{zhang2025icedit,
  title = {{Enabling Instructional Image Editing with In-Context Generation in Large Scale Diffusion Transformer}},
  author = {Zhang, Zechuan and Xie, Ji and Lu, Yu and Yang, Zongxin and Yang, Yi},
  booktitle = {Proceedings of the NeurIPS},
  year = {2025}
}

@inproceedings{mao2026varedit,
  title = {{Visual Autoregressive Modeling for Instruction-Guided Image Editing}},
  author = {Mao, Qingyang and Cai, Qi and Li, Yehao and Pan, Yingwei and Cheng, Mingyue and Yao, Ting and Liu, Qi and Mei, Tao},
  booktitle = {Proceedings of the ICLR},
  year = {2026}
}

@inproceedings{lu2026chordedit,
  title = {{ChordEdit: One-Step Low-Energy Transport for Image Editing}},
  author = {Lu, Liangsi and Chen, Xuhang and Guo, Minzhe and Li, Shichu and Wang, Jingchao and Shi, Yang},
  booktitle = {Proceedings of the CVPR},
  year = {2026}
}

@inproceedings{mao2025aceplus,
  title = {{ACE++: Instruction-Based Image Creation and Editing via Context-Aware Content Filling}},
  author = {Mao, Chaojie and Zhang, Jingfeng and Pan, Yulin and Jiang, Zeyinzi and Han, Zhen and Liu, Yu and Zhou, Jingren},
  booktitle = {Proceedings of the ICCV Workshops},
  year = {2025}
}

@article{labs2025flux,
  title={FLUX.1 Kontext: Flow Matching for In-Context Image Generation and Editing in Latent Space},
  author={{Black Forest Labs} and Batifol, Stephen and Blattmann, Andreas and Boesel, Frederic and Consul, Saksham and Diagne, Cyril and Dockhorn, Tim and English, Jack and English, Zion and Esser, Patrick and others},
  journal={arXiv preprint arXiv:2506.15742},
  year={2025}
}

@inproceedings{yang2026videocof,
  title = {{VideoCoF: Unified Video Editing with Temporal Reasoner}},
  author = {Yang, Xiangpeng and Xie, Ji and Yang, Yiyuan and Ma, Yue and Huang, Yan and Xu, Min and Wu, Qiang},
  booktitle = {Proceedings of the CVPR},
  year = {2026}
}

@article{yang2026omni,
  title={Omni-Video 2: Scaling MLLM-Conditioned Diffusion for Unified Video Generation and Editing},
  author={Yang, Hao and Tan, Zhiyu and Gong, Jia and Qin, Luozheng and Chen, Hesen and Yang, Xiaomeng and Sun, Yuqing and Lin, Yuetan and Yang, Mengping and Li, Hao},
  journal={arXiv preprint arXiv:2602.08820},
  year={2026}
}

@inproceedings{jiang2025vace,
  title = {{VACE: All-in-One Video Creation and Editing}},
  author = {Jiang, Zeyinzi and Han, Zhen and Mao, Chaojie and Zhang, Jingfeng and Pan, Yulin and Liu, Yu},
  booktitle = {Proceedings of the ICCV},
  year = {2025}
}

@article{zhou2018stereo,
  title = {{Stereo Magnification: Learning View Synthesis using Multiplane Images}},
  author = {Zhou, Tinghui and Tucker, Richard and Flynn, John and Fyffe, Graham and Snavely, Noah},
  journal = {ACM Transactions on Graphics},
  year = {2018}
}

@article{zheng2025realcam,
  title={RealCam-Vid: High-resolution video dataset with dynamic scenes and metric-scale camera movements},
  author={Zheng, Guangcong and Li, Teng and Zhou, Xianpan and Li, Xi},
  journal={arXiv preprint arXiv:2504.08212},
  year={2025}
}

@article{bai2025qwen3vl,
  title   = {{Qwen3-VL} Technical Report},
  author  = {Bai, Shuai and Cai, Yuxuan and Chen, Ruizhe and Chen, Keqin and
             Chen, Xionghui and Cheng, Zesen and Deng, Lianghao and Ding, Wei and
             Gao, Chang and Ge, Chunjiang and others},
  journal = {arXiv preprint arXiv:2511.21631},
  year    = {2025}
}

@misc{qwen2026qwen35,
  title = {{Qwen3.5: Towards Native Multimodal Agents}},
  author = {{Qwen Team}},
  year = {2026}
}

@article{team2025kling,
  title={Kling-Omni Technical Report},
  author={{Kling Team} and Chen, Jialu and Ci, Yuanzheng and Du, Xiangyu and Feng, Zipeng and Gai, Kun and Guo, Sainan and Han, Feng and He, Jingbin and He, Kang and others},
  journal={arXiv preprint arXiv:2512.16776},
  year={2025}
}

@inproceedings{ju2024pnp,
  title={{PnP Inversion: Boosting Diffusion-based Editing with 3 Lines of Code}},
  author={Ju, Xuan and Zeng, Ailing and Bian, Yuxuan and Liu, Shaoteng and Xu, Qiang},
  booktitle={Proceedings of the ICLR},
  year={2024}
}

@inproceedings{pan2025icebench,
  title={{ICE-Bench: A Unified and Comprehensive Benchmark for Image Creating and Editing}},
  author={Pan, Yulin and He, Xiangteng and Mao, Chaojie and Han, Zhen and Jiang, Zeyinzi and Zhang, Jingfeng and Liu, Yu},
  booktitle={Proceedings of the ICCV},
  year={2025}
}

@inproceedings{li2025fivebench,
  title={{FiVE-Bench: A Fine-grained Video Editing Benchmark for Evaluating Emerging Diffusion and Rectified Flow Models}},
  author={Li, Minghan and Xie, Chenxi and Wu, Yichen and Zhang, Lei and Wang, Mengyu},
  booktitle={Proceedings of the ICCV},
  year={2025}
}

@inproceedings{wu2023dover,
  title={{Exploring Video Quality Assessment on User Generated Contents from Aesthetic and Technical Perspectives}},
  author={Wu, Haoning and Zhang, Erli and Liao, Liang and Chen, Chaofeng and Hou, Jingwen and Wang, Annan and Sun, Wenxiu and Yan, Qiong and Lin, Weisi},
  booktitle={Proceedings of the ICCV},
  year={2023}
}

@article{ravi2024sam2,
  title={{SAM 2: Segment Anything in Images and Videos}},
  author={Ravi, Nikhila and Gabeur, Valentin and Hu, Yuan-Ting and Hu, Ronghang and Ryali, Chaitanya and Ma, Tengyu and Khedr, Haitham and R{\"a}dle, Roman and Rolland, Chloe and Gustafson, Laura and Mintun, Eric and Pan, Junting and Alwala, Kalyan Vasudev and Carion, Nicolas and Wu, Chao-Yuan and Girshick, Ross and Doll{\'a}r, Piotr and Feichtenhofer, Christoph},
  journal={arXiv preprint arXiv:2408.00714},
  year={2024}
}

@inproceedings{karaev2024cotracker,
  title={{CoTracker: It is Better to Track Together}},
  author={Karaev, Nikita and Rocco, Ignacio and Graham, Benjamin and Neverova, Natalia and Vedaldi, Andrea and Rupprecht, Christian},
  booktitle={Proceedings of the ECCV},
  year={2024}
}

@inproceedings{barron2022mipnerf360,
  title={{Mip-NeRF 360: Unbounded Anti-Aliased Neural Radiance Fields}},
  author={Barron, Jonathan T. and Mildenhall, Ben and Verbin, Dor and Srinivasan, Pratul P. and Hedman, Peter},
  booktitle={Proceedings of the CVPR},
  year={2022}
}

@inproceedings{yao2020blendedmvs,
  title={{BlendedMVS: A Large-Scale Dataset for Generalized Multi-View Stereo Networks}},
  author={Yao, Yao and Luo, Zixin and Li, Shiwei and Zhang, Jingyang and Ren, Yufan and Zhou, Lei and Fang, Tian and Quan, Long},
  booktitle={Proceedings of the CVPR},
  year={2020}
}

@article{kerbl2023gaussians,
  title={{3D Gaussian Splatting for Real-Time Radiance Field Rendering}},
  author={Kerbl, Bernhard and Kopanas, Georgios and Leimk{\"u}hler, Thomas and Drettakis, George},
  journal={ACM Transactions on Graphics},
  year={2023}
}

@inproceedings{wang2025vggt,
  title={{VGGT: Visual Geometry Grounded Transformer}},
  author={Wang, Jianyuan and Chen, Minghao and Karaev, Nikita and Vedaldi, Andrea and Rupprecht, Christian and Novotny, David},
  booktitle={Proceedings of the CVPR},
  year={2025}
}

@inproceedings{hu2022lora,
  title = {{LoRA: Low-Rank Adaptation of Large Language Models}},
  author = {Hu, Edward J. and Shen, Yelong and Wallis, Phillip and Allen-Zhu, Zeyuan and Li, Yuanzhi and Wang, Shean and Wang, Lu and Chen, Weizhu},
  booktitle = {Proceedings of the ICLR},
  year = {2022}
}
